\documentclass{article}

\PassOptionsToPackage{numbers, compress}{natbib}

\usepackage[utf8]{inputenc} %
\usepackage[T1]{fontenc}    %
\usepackage{hyperref}       %
\usepackage{url}            %
\usepackage{booktabs}       %
\usepackage{amsfonts}       %
\usepackage{nicefrac}       %
\usepackage{microtype}      %
\usepackage{natbib}
\usepackage{ mathrsfs }
\usepackage[pdftex]{graphicx}
\usepackage{mathtools}
\usepackage{subcaption}
\usepackage{realboxes}
\usepackage{tabto}
\usepackage{helvet}
\usepackage{tabularx}
\usepackage{rotating}
\usepackage{algorithmic}
\usepackage[title]{appendix}

\usepackage{geometry}
\newsavebox{\fmbox}
\newenvironment{fmpage}[1]
{\begin{lrbox}{\fmbox}\begin{minipage}{#1}}
{\end{minipage}\end{lrbox}\fbox{\usebox{\fmbox}}}

\title{ \textbf{Empirical Explorations in Training Networks with\\Discrete Activations} }

\author{
  Shumeet Baluja \\
  Google, Inc. \\
  \texttt{shumeet@google.com} \\
}

\begin{document}

\maketitle

\begin{abstract}

  We present extensive experiments training and testing hidden units
  in deep networks that emit only a predefined, static, number of
  discretized values.  These units provide benefits in real-world
  deployment in systems in which memory and/or computation may be
  limited.  Additionally, they are particularly well suited for use in
  large recurrent network models that require the maintenance of
  large amounts of internal state in memory.  Surprisingly, we find
  that despite reducing the number of values that can be represented
  in the output activations from $2^{32}-2^{64}$ to between 64 and
  256, there is little to no degradation in network performance across
  a variety of different settings.  We investigate simple
  classification and regression tasks, as well as memorization and
  compression problems.  We compare the results with more standard
  activations, such as tanh and relu.  Unlike previous discretization
  studies which often concentrate only on binary units, we examine the
  effects of varying the number of allowed activation levels.  Compared to existing approaches for
  discretization, the approach presented here is both conceptually and
  programatically simple, has no stochastic component, and allows the
  training, testing, and usage phases to be treated in exactly the
  same manner.
  
\end{abstract}

\section{Introduction and Related Work}

Almost all popular neural network training algorithms rely on
gradient-based learning. For reliable computation of the gradients, it
is useful for the hidden unit activations to be continuous and smooth.
If the activation has large plateaus or discontinuities,
gradient-based learning becomes difficult or even impossible.  This is
a large part of what motivated the move from neural networks based on
synthetic discrete neurons, with hard thresholds, to the use of
units with a sigmoid non-linearity, as well as the back-propagation
algorithm to compute the
gradients~\cite{rumelhart1988learning,bengio2013estimating}.
Interestingly, despite the use of the sigmoid non-linearity to smooth
the derivatives, even early on in neural network research it was
observed that often, after training, the units clustered their
activations around the extrema~\cite{weigend1994overfitting} --
thereby potentially under-utilizing their full representational
capacity.

Recently, there has been renewed interest in using discrete outputs
for the activation of the hidden units and weights of a trained network.
Though perhaps closer in some regards to biologically plausible
spiking neurons, much of the research in discretization of outputs and
weights has stemmed from pragmatic concerns.  These units provide
benefits in the deployment of systems in which memory and/or
computation may be limited, such as cell-phones and specialized
hardware designed for the forward propagation of large networks.
Additionally, they are particularly well suited for use in large
recurrent network models that require the maintenance of large
amounts of internal state in memory.

To date, most commonly, research has focused on ``binarizing'' the
networks -- both the weights and activations (larger alphabets are
less often explored, though work has been done in that direction;
see~\cite{li2016}).  The focus of this paper is on the discretization
of the outputs of the hidden units.  Unlike previous studies, we will
empirically examine what happens when the units are allowed to output
between 2 to 256 discrete values.  The trade-offs between higher
cardinality alphabets and the number of employed hidden units will be
examined in the experiments.

In the last decade, the use of rectified linear units, as well as a
number of other non-linearities, has shown that it is possible to
train networks that do not strictly adhere to the smoothness
constraints often considered as necessary
~\cite{goodfellow2013maxout,glorot2011deep,nair2010rectified}-- even
with only few or no changes to the learning algorithms.  When purely
discrete outputs are desired, however, such as with binary units, a
number of additional steps are normally
taken~\cite{raiko2014techniques,bengio2013estimating,hou2016loss,courbariaux2016binarized,tang2013learning,maddison2016} or evolutionary strategies used~\cite{plagianakos2001training}.
At a high level, many of the methods employ a stochastic binary unit
and inject noise during the forward pass to sample
the units and the associated effect on the network's outputs.  With
this estimation, it is possible to calculate a gradient and pass it
through the network.  One interesting benefit of this
method is its use in generative networks. In generative networks, if the units are
employed stochastically in the forward propagation phase, they can be beneficial for generating
multiple responses to a single input.  For example, see~\cite{raiko2014techniques} (in particular the task of generating the
bottom half of hand written digits, given only the top-half). In the
same work,~\cite{raiko2014techniques} extend~\cite{tang2013learning}
to show that learning with stochastic units may not even be necessary
if they are used within a larger deterministic network.  Other
existing binarization methods (e.g.~\cite{courbariaux2016binarized})
liken the process to dropout~\cite{srivastava2014dropout} and its
regularization effects. Instead of randomly setting activations to
zero when computing gradients, they binarize both the activations and
the weights.

\section {Approach \& Intuition}

\begin{figure}
  \begin{center}

    \includegraphics[height=1in]{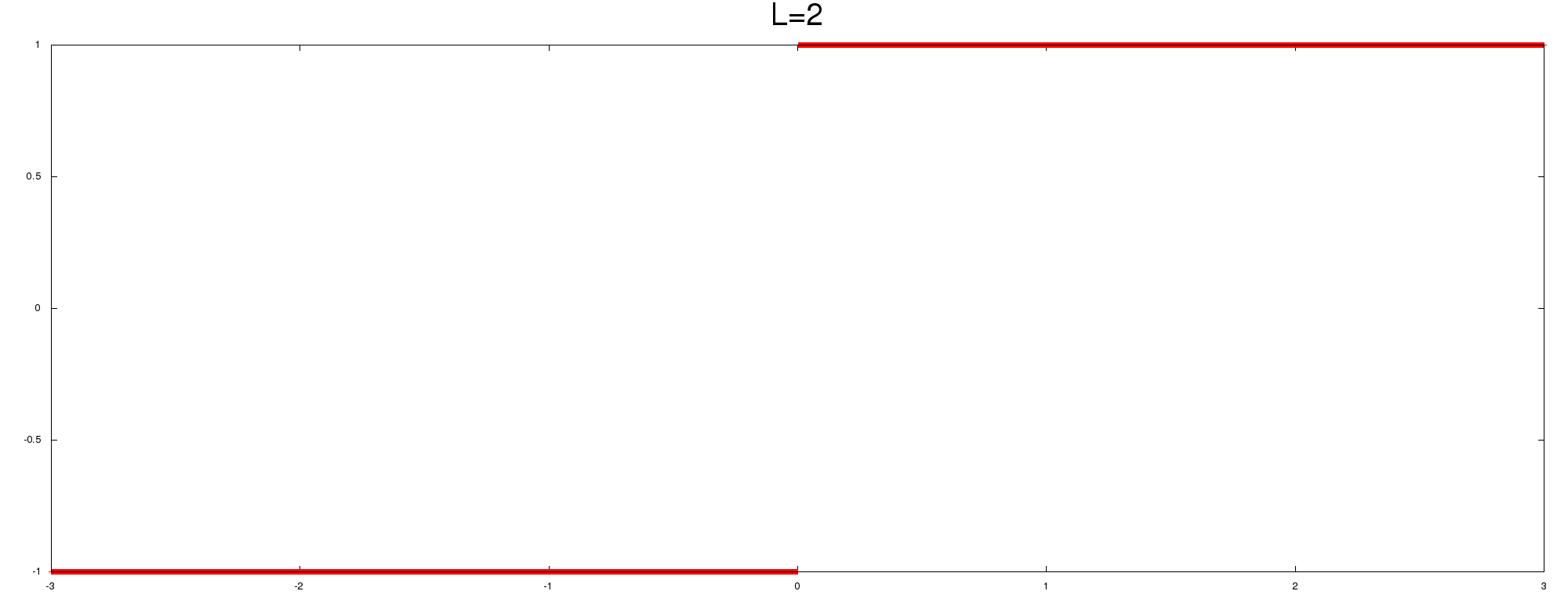}
    
    \includegraphics[height=1in]{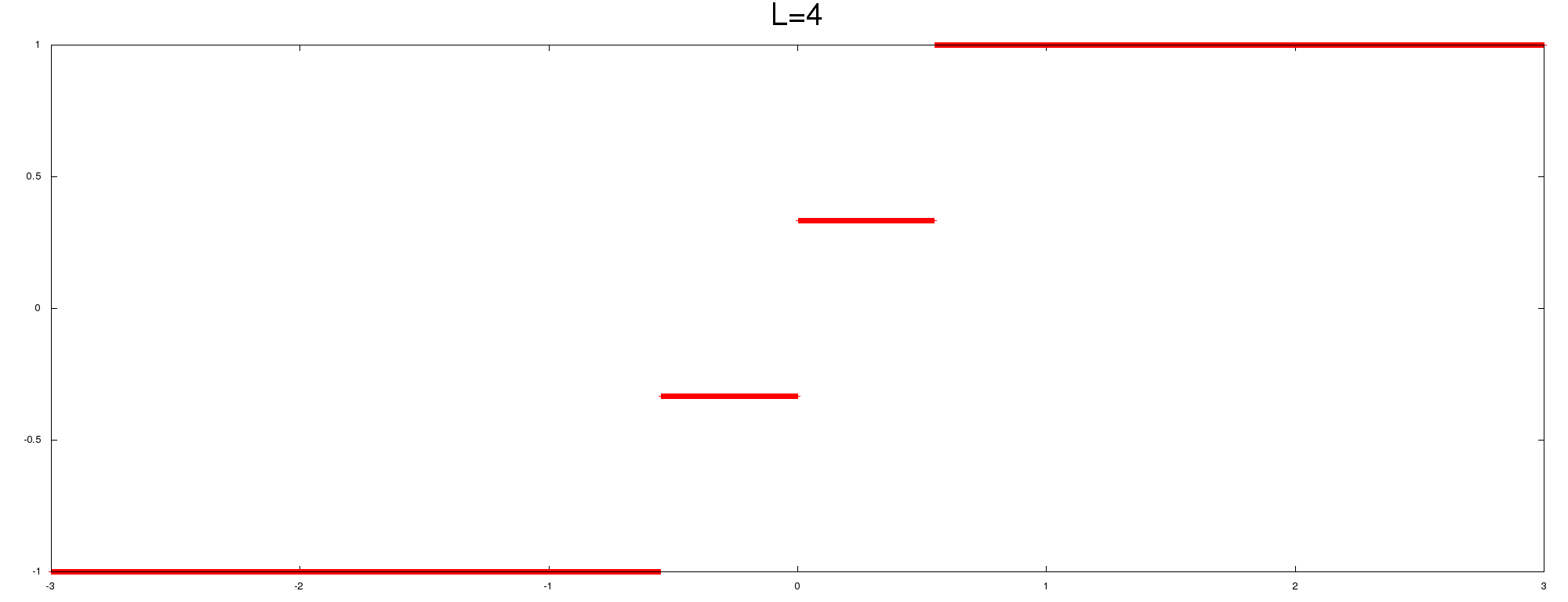}
    
    \includegraphics[height=1in]{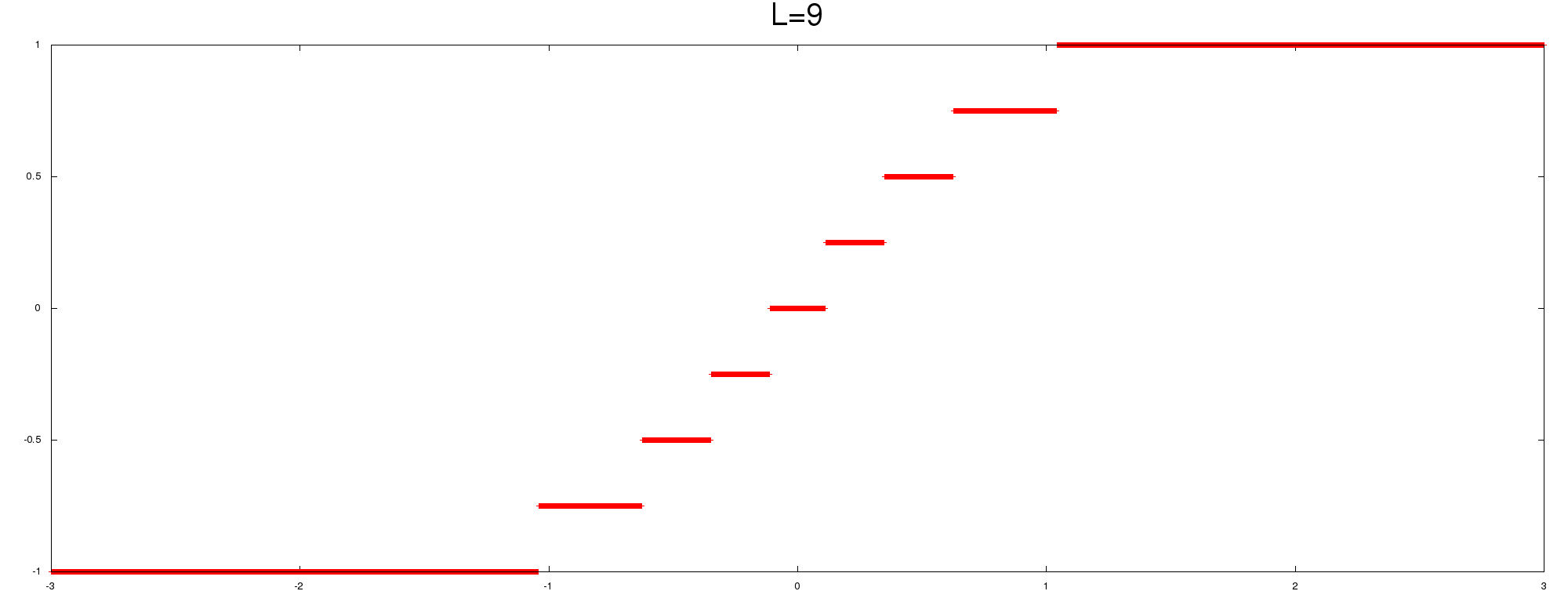}
    
    \includegraphics[height=1in]{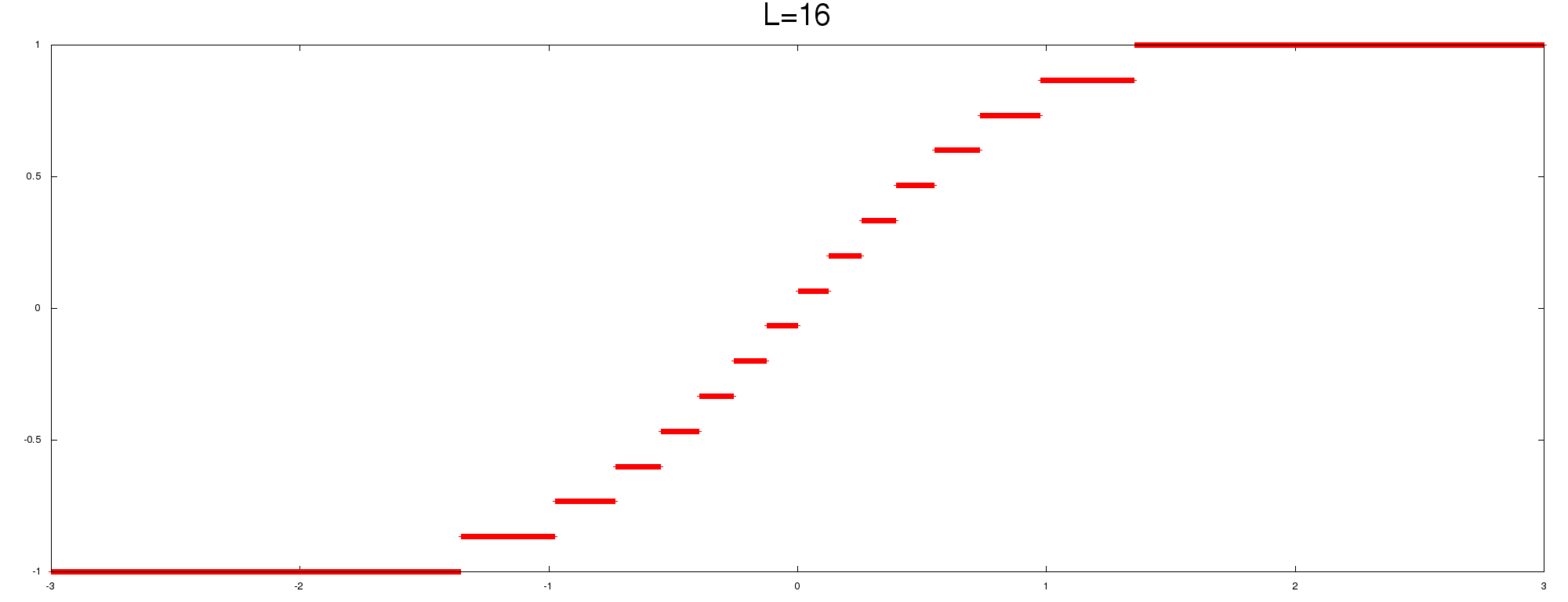}
    
    \includegraphics[height=1in]{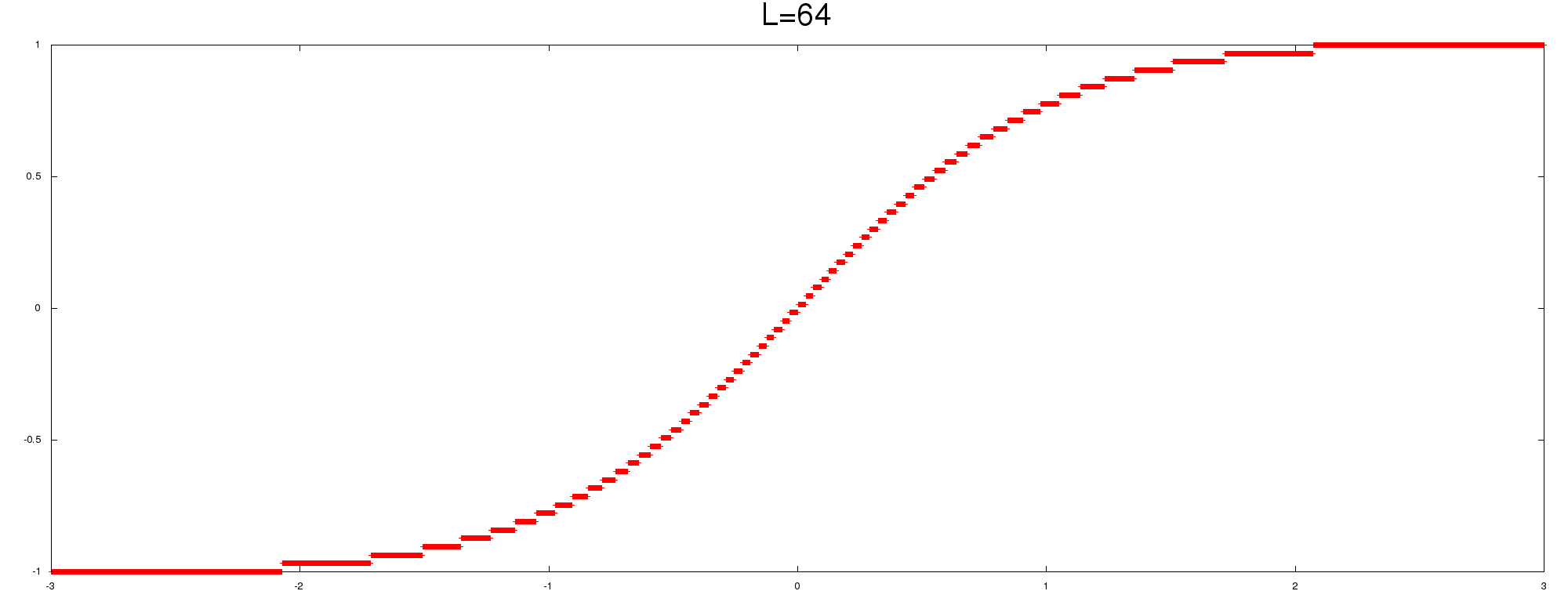}
    
    \includegraphics[height=1in]{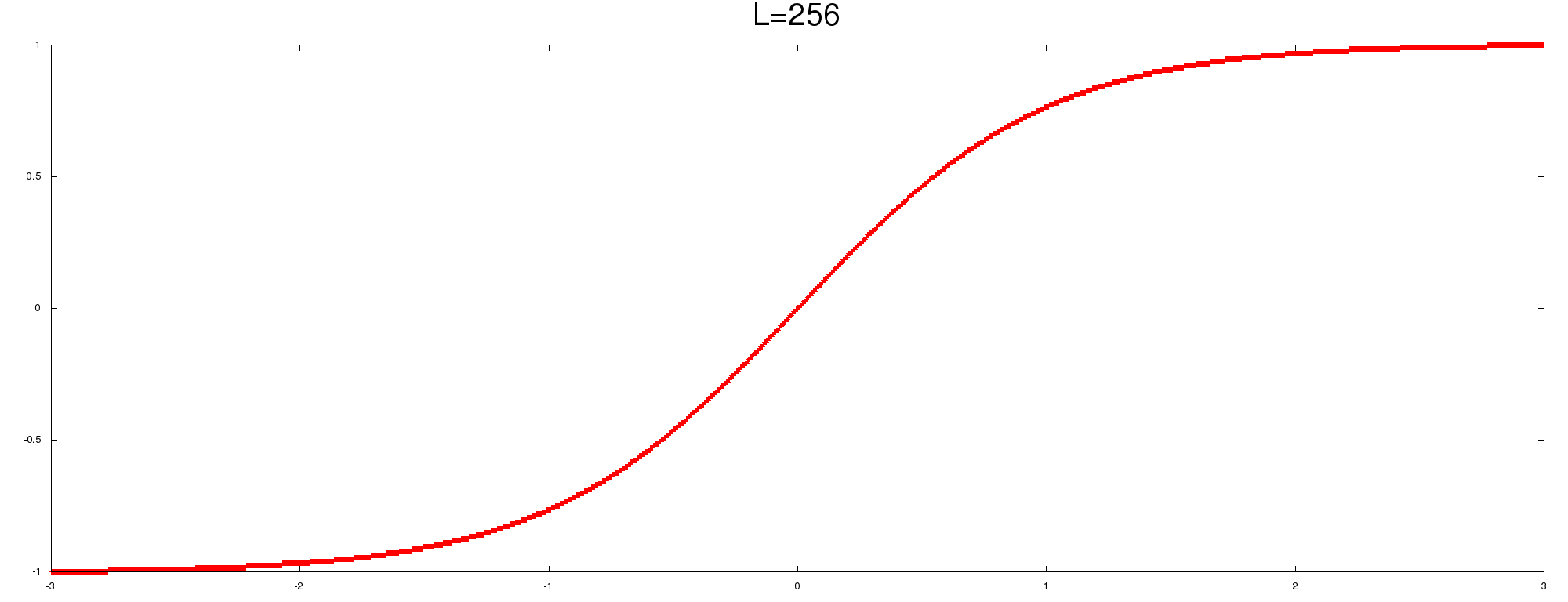}

\caption{SUDO units, shown with 2, 4, 9, 64 and 256 levels.  In the
  regions of largest slope in the underlying sigmoid/tanh function,
  the discretization levels change the fastest.  There is no
  requirement to constrain the number of discretization levels to a
  power of 2 (as shown with $L=9$), though in practice that may be
  preferred for memory efficiency.}
\label{fig:sudounits}
\end{center}
\end{figure}

In this section, we describe the SUDO
unit: \emph{S}igmoid-\emph{U}nderlying, \emph{D}iscrete-\emph{O}utput
Units.  In its simplest implementation, the SUDO unit is instantiated
with a pre-defined set of output discretization levels, $L$, that
output a value between between a bounded range -- e.g. either 0 and 1
or -1 and +1.  See Figure~\ref{fig:sudounits}.   When the output is
scaled between -1 and +1, the discretized output is computed as
follows (shown in expanded form for clarity):
\newlength\myindent %
\setlength\myindent{1em} %
\newcommand\bindent{%
  \begingroup %
  \setlength{\itemindent}{\myindent} %
  \addtolength{\algorithmicindent}{\myindent} %
}
\newcommand\eindent{\endgroup} %
\\
\vskip 0.25in
\begin{fmpage}{0.9\textwidth}

\begin{algorithmic}
    \STATE function SUDO\_Activation (input, levels):
    \bindent
    \STATE underlying  $\gets$ $tanh (input)$    
    \STATE activation\_step $\gets$ $2 / (levels - 1)$
    \STATE plateauRange    $\gets$ $2 / levels$

    \STATE output $\gets$ $(\left \lceil{ (underlying + 1.0) / plateauRange}\right \rceil- 1.0) * activation\_step$
    \STATE return ($-1.0 + output$)
     \eindent    
  \end{algorithmic}
\end{fmpage}
\vskip 0.25in

\begin{itemize}
\item \emph{On a practical note:} to replicate the results in this
  study or to use SUDO units within standard neural network packages,
  a potential unexpected behavior should be avoided.  If the $tanh$
  function is approximated as -1.0 for very small large negative
  values of $input$ (the lower extrema) this function, as shown, yields an extra
  discretization level.  This has been noted as a subtle problem in a
  popular language/package.  There are many simple ``tweaks'' that can
  be used to avoid this issue that do not change the effectiveness of
  the procedure -- such as clipping the outputs or multiplying the
  $tanh (input)$ by a number smaller than, but close to, 1.0 (e.g. 0.9999).

\end{itemize}

While there have been numerous studies that have examined the effects
of binarizing activations, as described in the previous section, there
have been fewer that have empirically examined the effects of
increasing the number of discretization levels, $L$.  As we will
demonstrate, as $L$ is increased, many of the difficulties of training
discretized units are mitigated, and simple mechanisms perform well.
This allows the use of all the currently popular training algorithms
\emph{with no modification}.

Of course, naively backpropagating errors with SUDO-units will quickly
run into problems as the activations are not well suited for
calculating derivatives: they are both discontinuous and are largely
characterized only by piece-wise constant functions.  In contrast,
standard sigmoid and tanh activations do not suffer from this problem.
Relu units do partially share the same difficulties; however, because
they are unbounded when positive, coupled with the fact that there is
only a single discontinuity, gradient based methods continue to work.

In order to use gradient based methods with SUDO units, we need to
define a suitable derivative.   We simply use the derivative of the
underlying function that we are discretizing.  For the case shown
above, the derivative is simply the derivative of $tanh(x)$ which is $
1.0 - tanh^2(x)$.   In the forward pass of the network, the output of
each unit is the discretized output.  In the backward pass, we simply
ignore the discretization and use the derivatives from the underlying
function.

Why could this work in training?  There are two important facets of
this activation function to consider.  First, if we had tried to use
the discretized outputs in the backpropagation phase, the plateaus
would not have given usable derivatives.  By ignoring the
discretizations, the weights of the network \emph{will still move in
  the wanted directions}.  The difference is, however, that unlike
with standard tanh units, any single move \emph{may not} actually change the
unit's output.  In an extreme case, it is possible that with a low
enough learning rate, the entire network's output may not change
despite all the weight changes made.
This extreme case may cause a slow down in training, but, most
importantly, it \emph{will not} end training.  Instead, in the next
backprop phase, the weights will again be directed to move, and of
those that move in the same direction, some will cross a
discretization threshold. This changes the unit's, and eventually,
the network's, output.%

Second, we need to carefully examine Figure~\ref{fig:sudounits}.  Note
that the plateaus are not evenly sized.  This is most easily
noticed in the middle two plots.  Note that where the magnitude of the
derivative for the underlying tanh function is maximum is where the
plateau is the smallest size.  This can be beneficial in practice; the
unit's output will change most rapidly where the derivative of tanh
changes the most rapidly, thereby keeping the expected movement closer
to the real movement through the discretization.

\begin{figure}
  \begin{center}

    \begin{tabular}{m{1em}VV|VVVVV}
      & tanh & relu &  SUDO-2 & SUDO-4 & SUDO-8 & SUDO-128 & SUDO-256  \\
      \midrule
    \centering \rotatebox{90}{$\leftarrow$ training epochs }&
    \includegraphics[height=3.22in,width=4.5em]{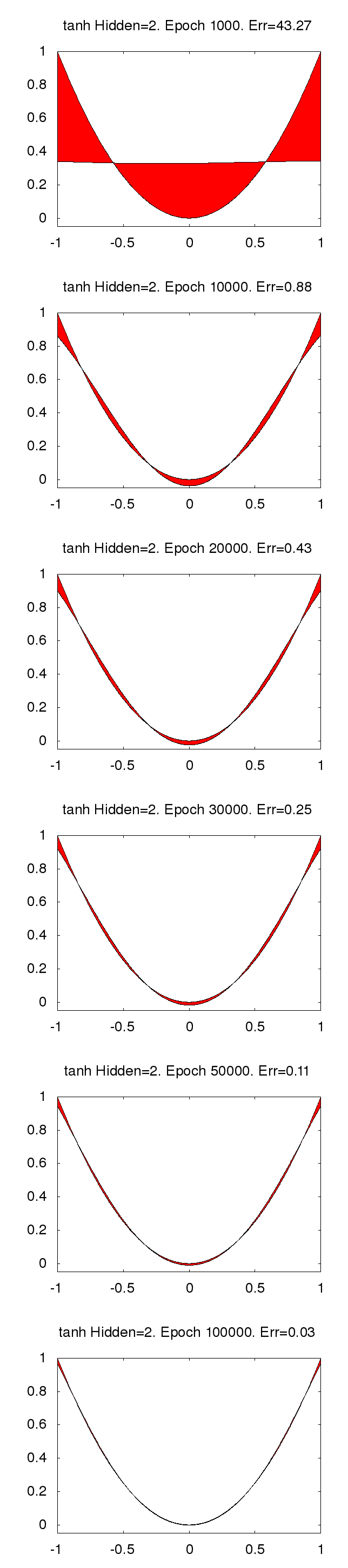}&
    \includegraphics[height=3.22in,width=4.5em]{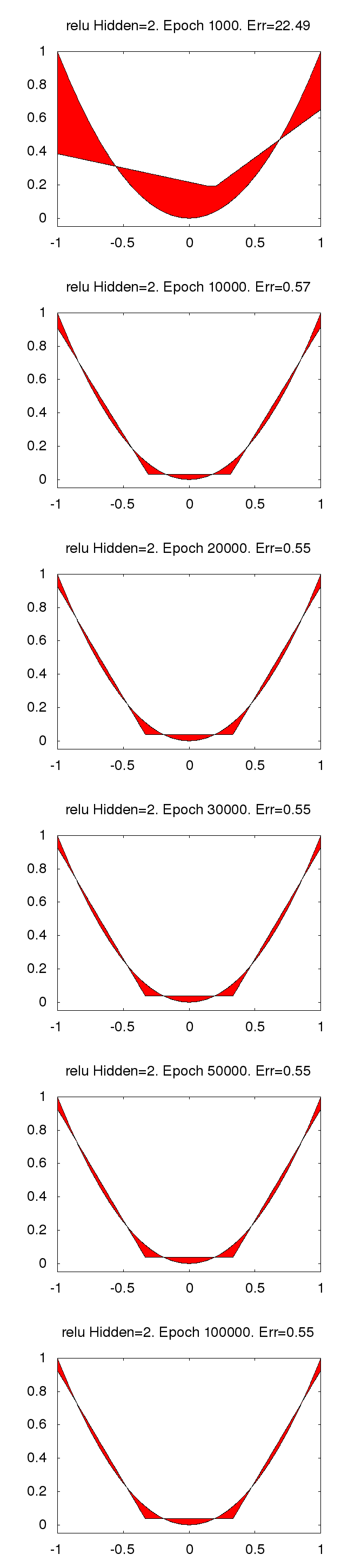}&
    \includegraphics[height=3.22in,width=4.5em]{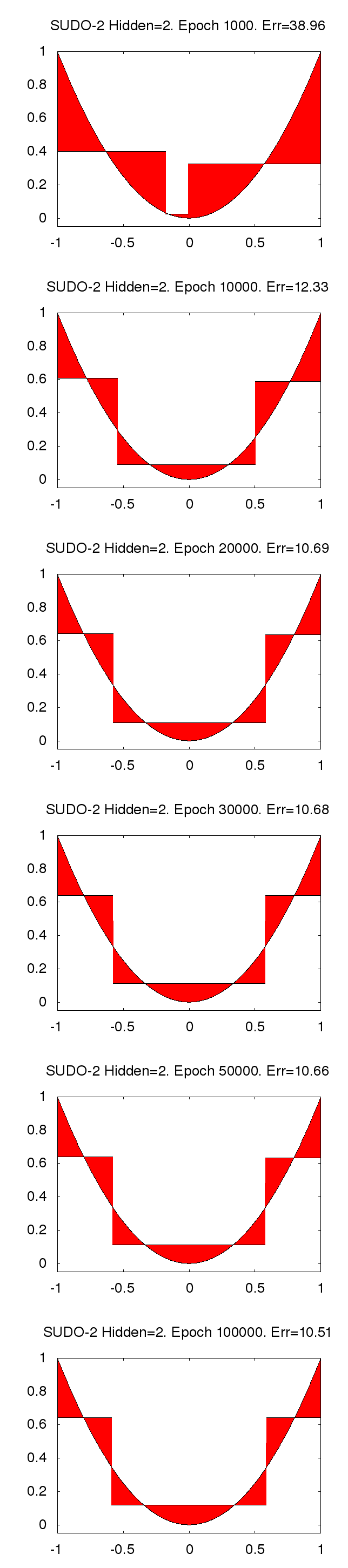}&
    \includegraphics[height=3.22in,width=4.5em]{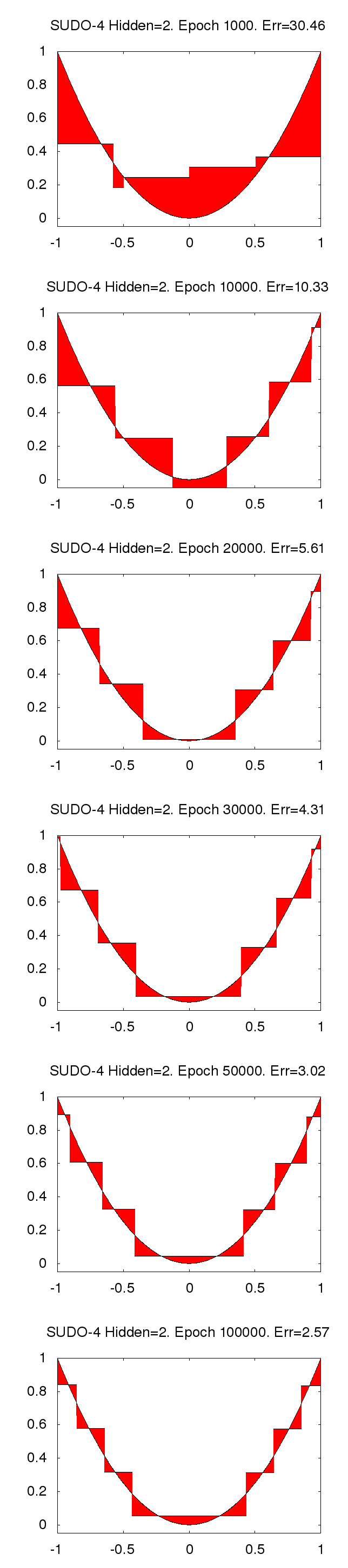}&
    \includegraphics[height=3.22in,width=4.5em]{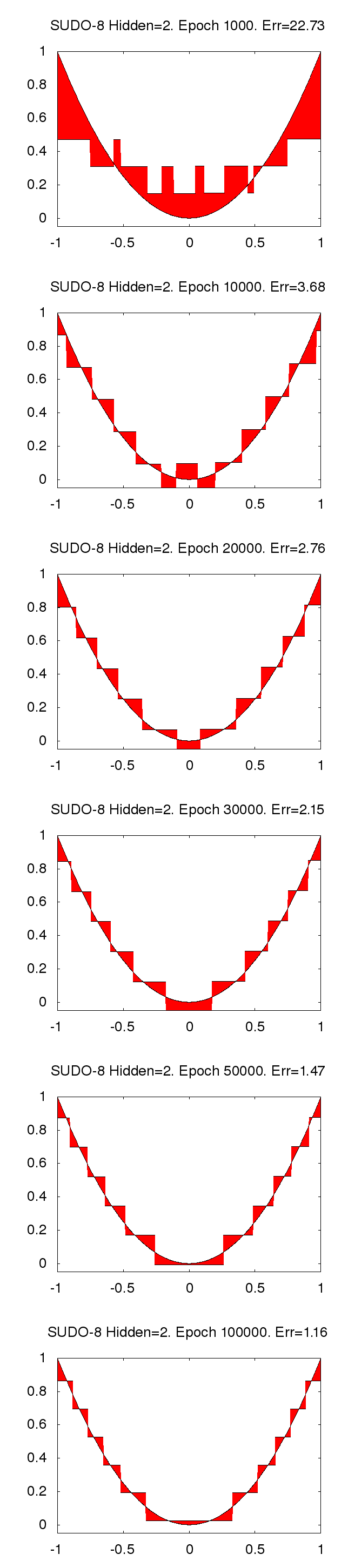}&
    \includegraphics[height=3.22in,width=4.5em]{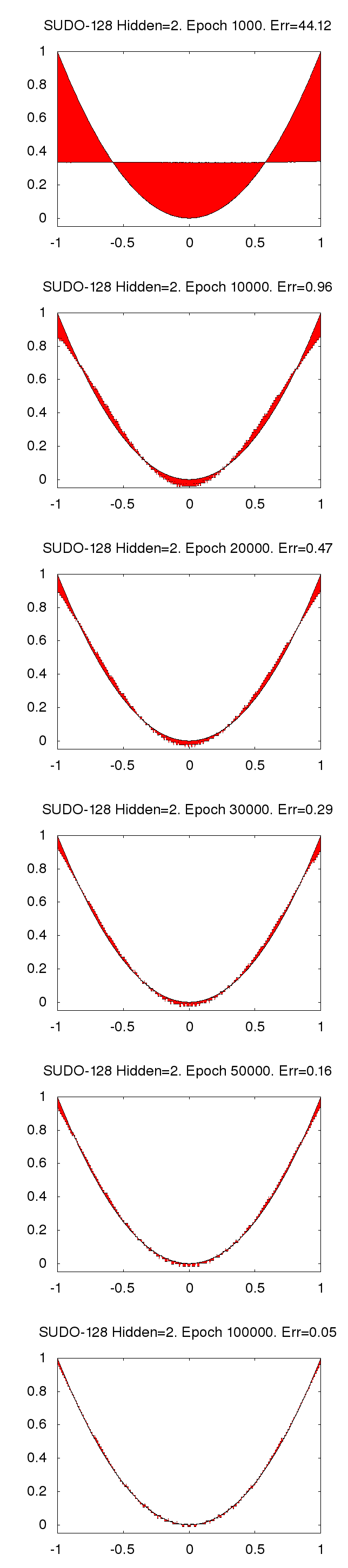}&
    \includegraphics[height=3.22in,width=4.5em]{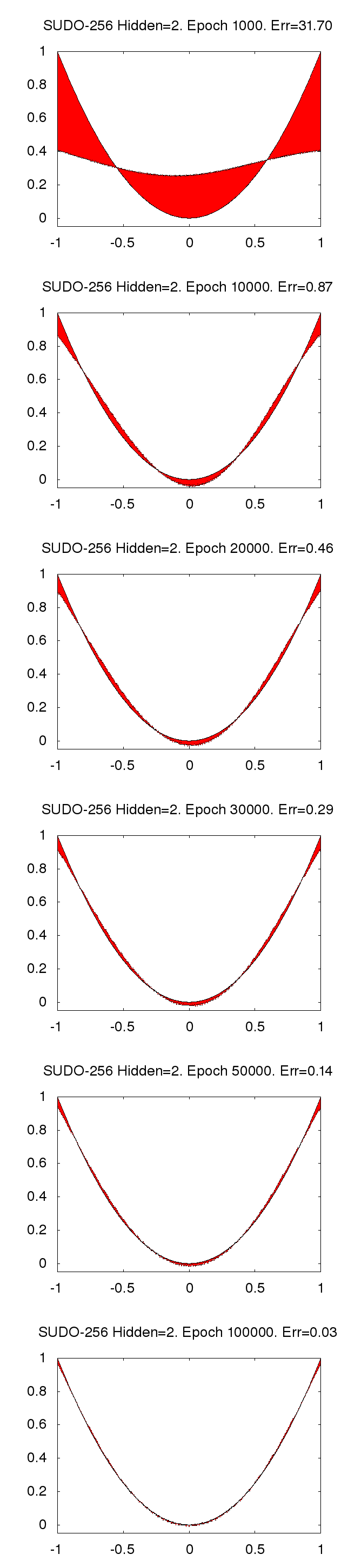}\\

    \end{tabular}

\caption{Training to fit a parabola with 2 hidden units. Area is the
  error between the actual and predicted.   Shown with
  hidden unit activations of (From Left to Right) Tanh, Relu, SUDO-2, SUDO-4, SUDO-8,
  SUDO-128 and SUDO-256.    From Top to Bottom:  Progress through epochs.  This provides an intuitive demonstration for how the discretized units change the network's performance.  In the most clear example, with SUDO-2, we see that the effect of the binary discretization levels in approximating the parabola.   The network has found a reasonable, symmetric, approximation, but cannot overcome the discretization artifacts -- with only the 2 hidden units. }
\label{fig:parabola2units}
\end{center}

\end{figure}

To further our intuition of how these units perform in practice, we
present three figures showing how SUDO, tanh and relu perform in tiny
networks trained to fit a one dimensional parabola.  The first,
Figure~\ref{fig:parabola2units}, shows how well the parabola is fit as
training progresses with a variety of activations.  For all of the
networks, there is a single linear output unit and two hidden units.
The network is trained with Stochastic Gradient Descent with momentum
(SGD+Momentum).

\begin{figure}
  \begin{center}
    
    \begin{tabular}{m{1em}VV|VVVVV}
      & tanh & relu &  SUDO-2 & SUDO-4 & SUDO-8 & SUDO-128 & SUDO-256  \\
      \midrule
    \centering \rotatebox{90}{$\leftarrow$ training epochs }&

    \includegraphics[height=3.22in,width=4.5em]{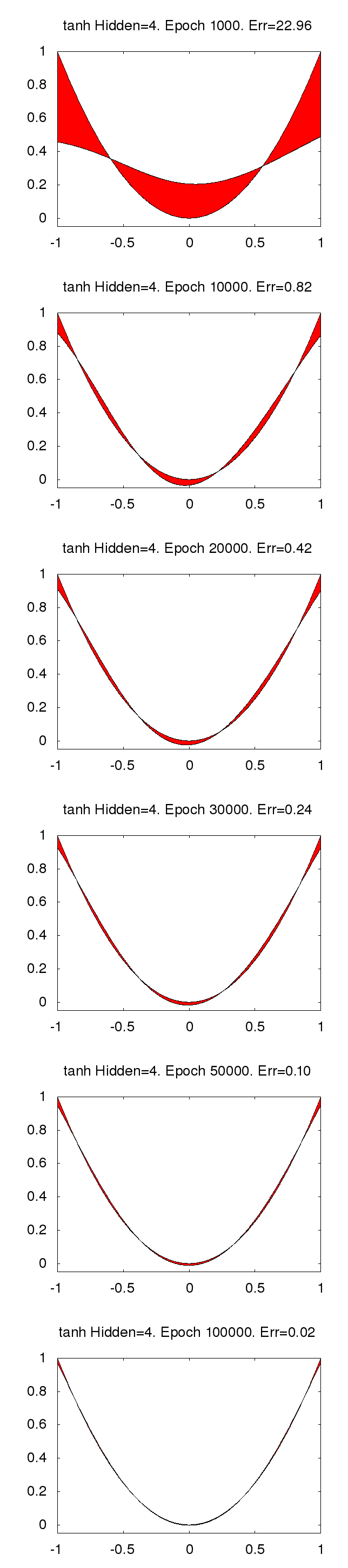}&
    \includegraphics[height=3.22in,width=4.5em]{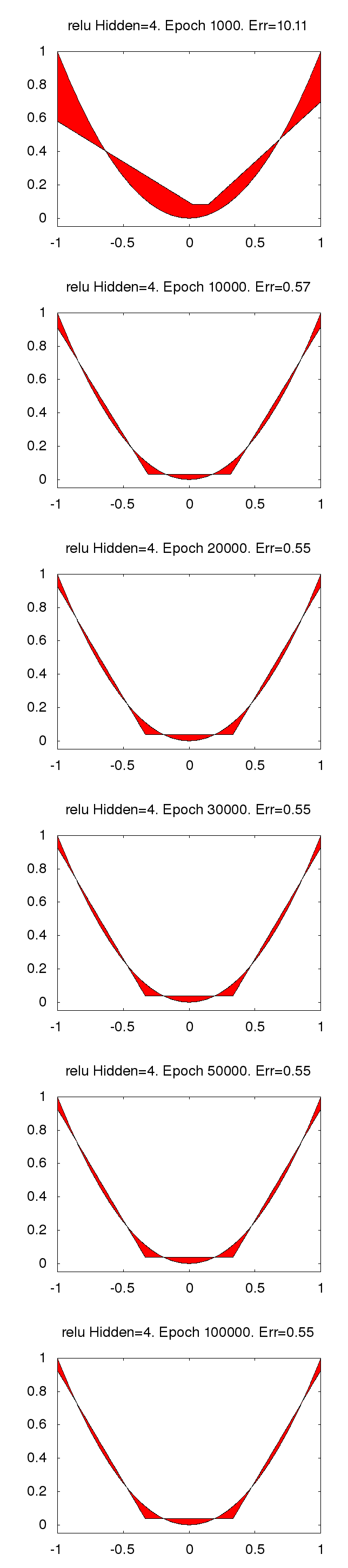}& 
    \includegraphics[height=3.22in,width=4.5em]{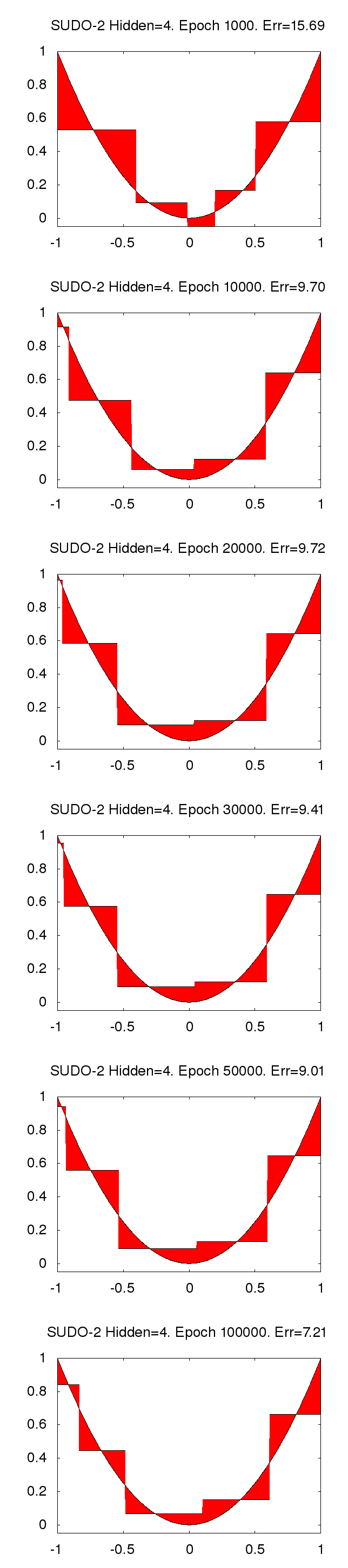}& 
    \includegraphics[height=3.22in,width=4.5em]{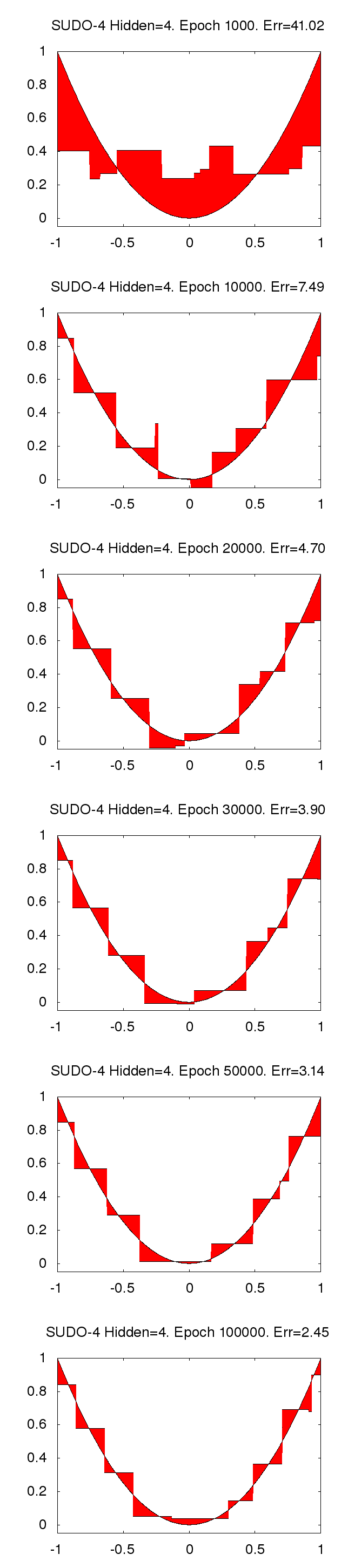}& 
    \includegraphics[height=3.22in,width=4.5em]{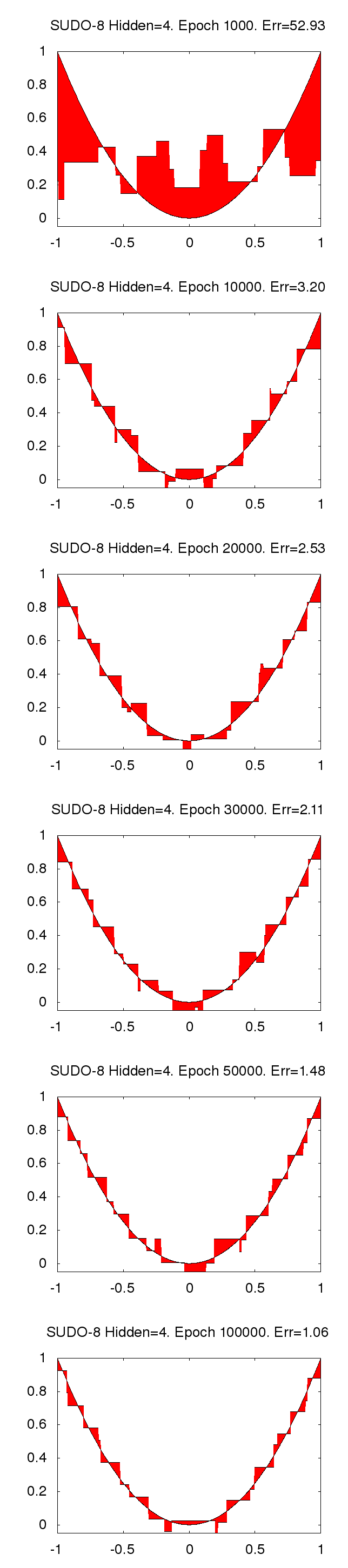}& 
    \includegraphics[height=3.22in,width=4.5em]{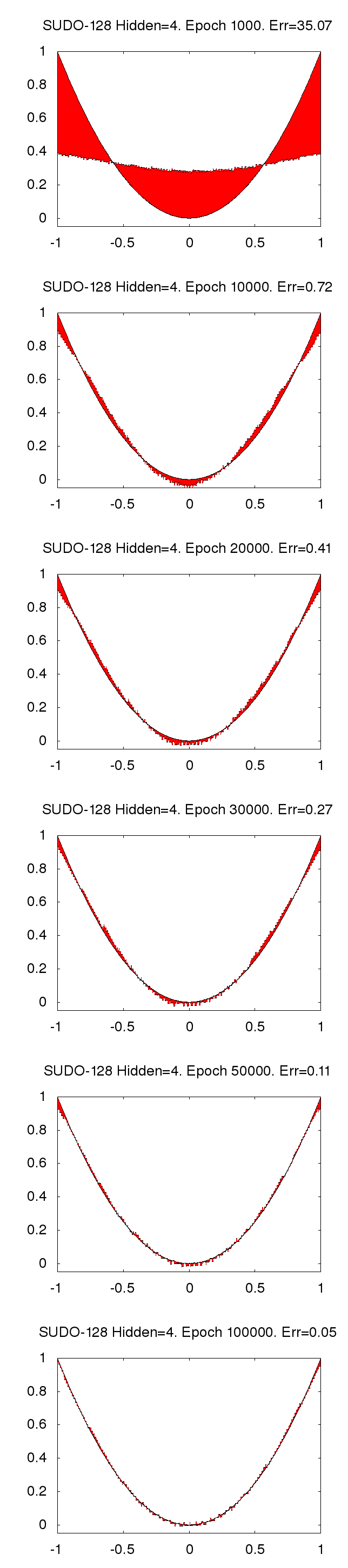}& 
    \includegraphics[height=3.22in,width=4.5em]{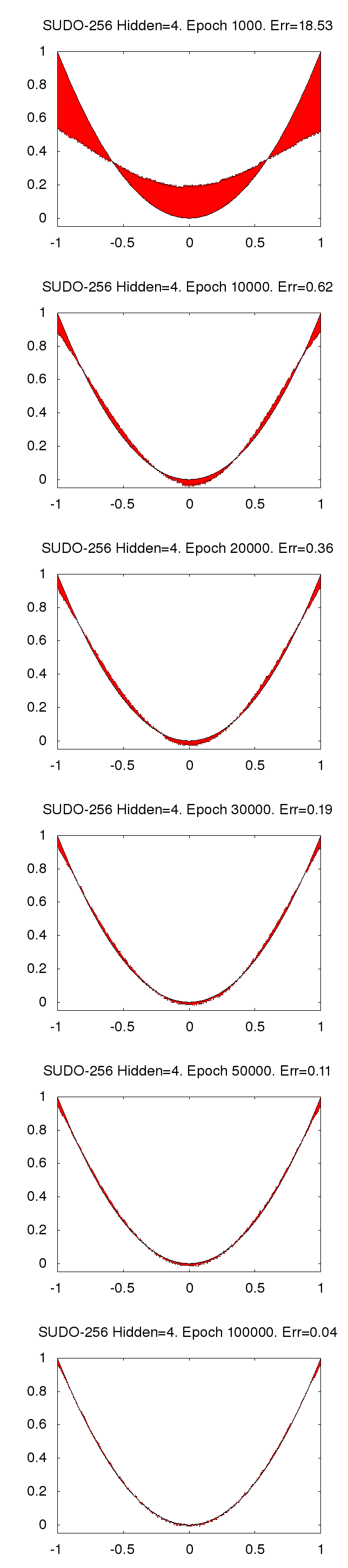}\\
    \end{tabular}

\caption{Training to fit a parabola with 4 hidden units. Area is the
  error between the actual and predicted.   Shown with
  hidden unit activations of (From Left to Right) Tanh, Relu, SUDO-2, SUDO-4, SUDO-8,
  SUDO-128 and SUDO-256.    From Top to Bottom:  Progress through
  epochs.}
\label{fig:parabola4units}
\end{center}

\end{figure}

\begin{figure}
  \begin{center}

    \begin{tabular}{m{1em}VV|VVVVV}
      & tanh & relu &  SUDO-2 & SUDO-4 & SUDO-8 & SUDO-128 & SUDO-256  \\
      \midrule
    \centering \rotatebox{90}{$\leftarrow$ training epochs }&

    \includegraphics[height=3.22in,width=4.5em]{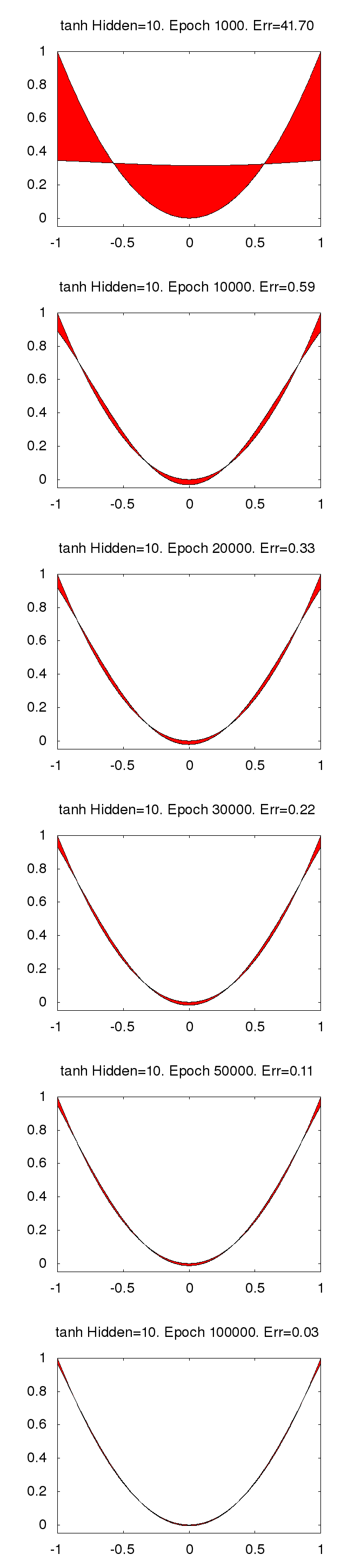}&
    \includegraphics[height=3.22in,width=4.5em]{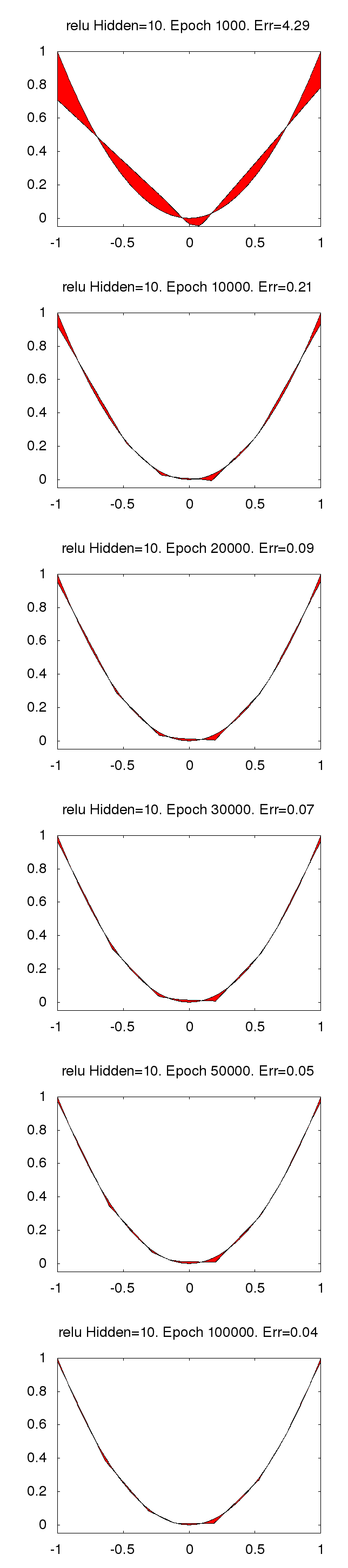}&
    \includegraphics[height=3.22in,width=4.5em]{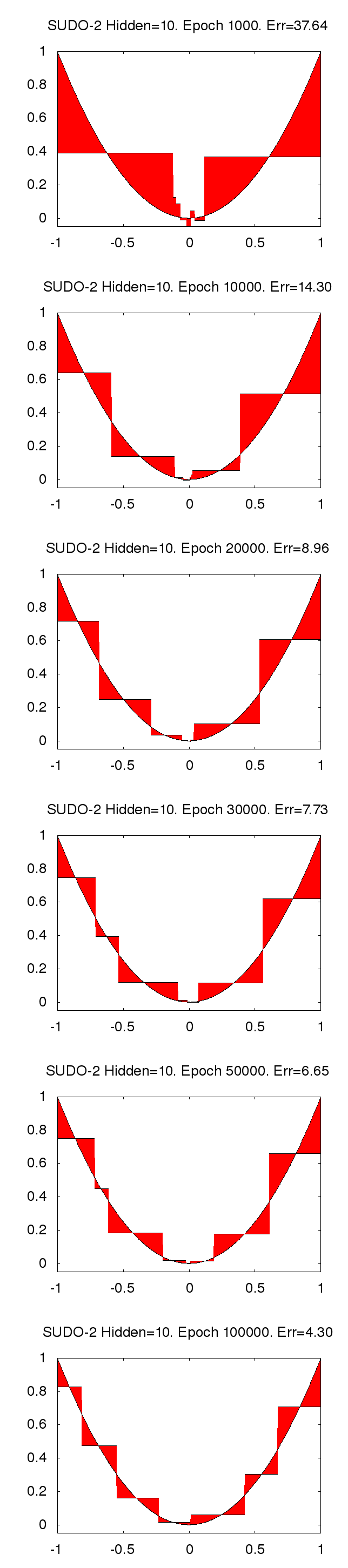}& 
    \includegraphics[height=3.22in,width=4.5em]{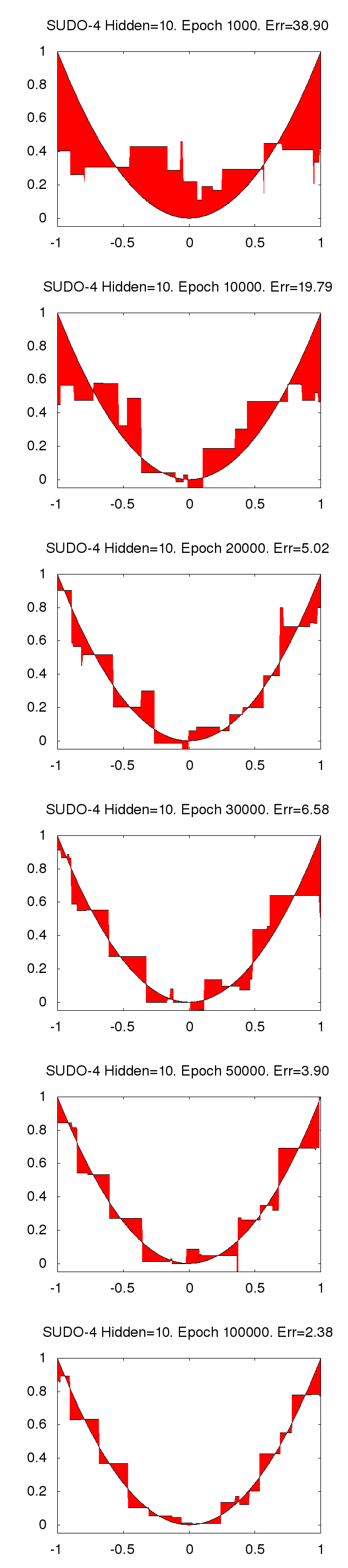}&
    \includegraphics[height=3.22in,width=4.5em]{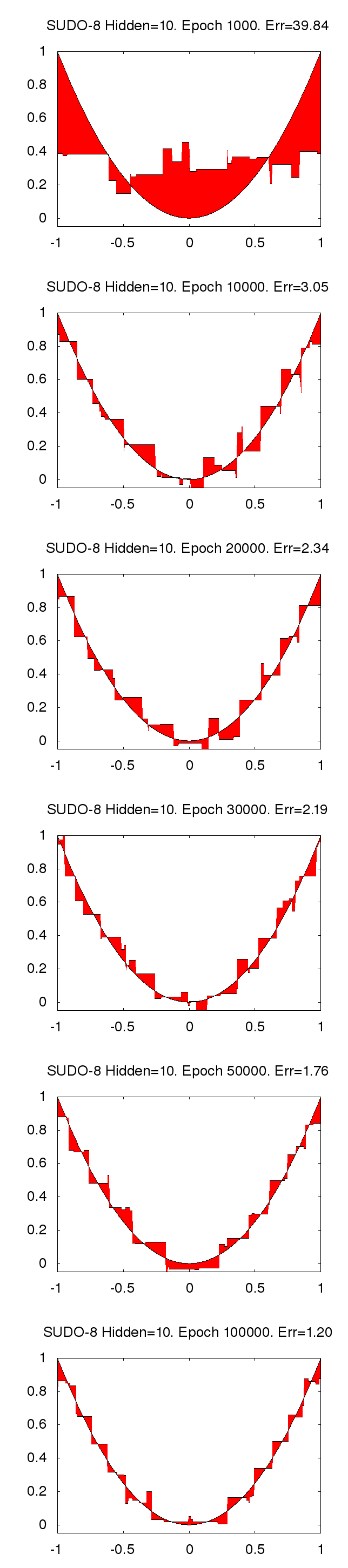}&
    \includegraphics[height=3.22in,width=4.5em]{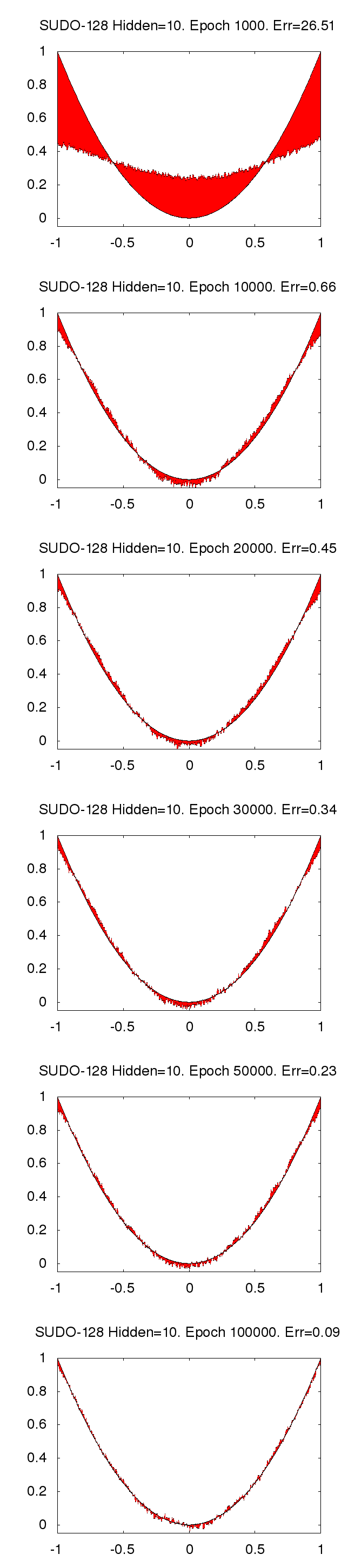}&
    \includegraphics[height=3.22in,width=4.5em]{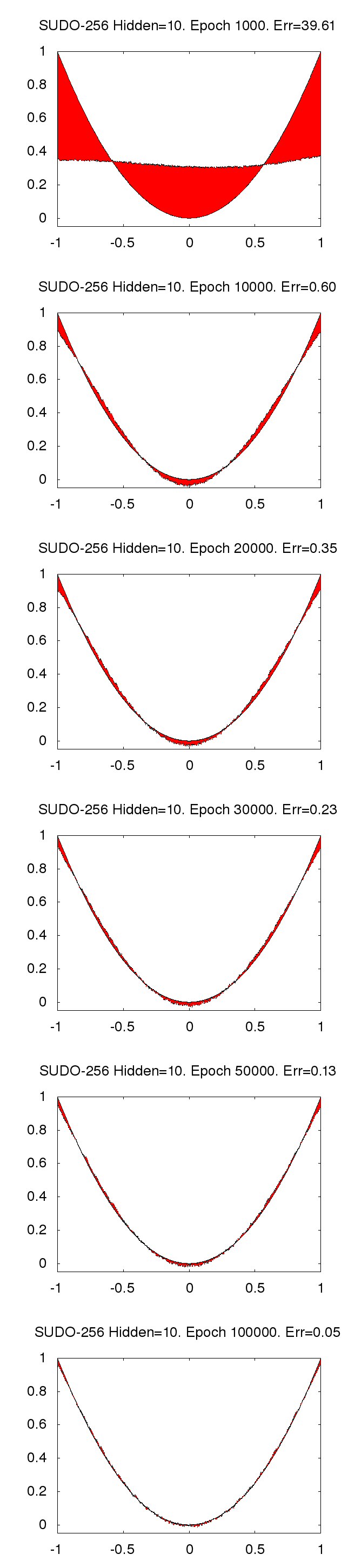}\\
    \end{tabular}

\caption{Training to fit a parabola with 10 hidden units. Area is the
  error between the actual and predicted.   Shown with
  hidden unit activations of (From Left to Right) Tanh, Relu, SUDO-2, SUDO-4, SUDO-8,
  SUDO-128 and SUDO-256.    From Top to Bottom:  Progress through
  epochs.}
\label{fig:parabola10units}
\end{center}

\end{figure}

Perhaps the most telling results are the training curves with SUDO
with $L=2$ (SUDO-2).  The resulting fit to the parabola matches
closely with intuition; the different levels of discretization
(almost) symmetrically reduce the error in a straightforward manner.
As $L$ is increased (moving to the right in the figure), the
performance matches the networks trained with tanh and relu
activations.  Figures~\ref{fig:parabola4units} and
~\ref{fig:parabola10units} show the same, but with networks that have
4 and 10 hidden units respectively.  Again, the performance follows close to
intuition.

\newcolumntype{V}{>{\centering\arraybackslash} m{4.5em} }

\clearpage
\section {Experiments}
\label {experiments}

In this section, we present a set of five experiments.  While modest
in scope in comparison to the large-scale vision tasks addressed by
recent deep-learning research, they serve to elucidate important
facets of the performance of the SUDO units. We explore a very large
number of architectures, network models, and hyper-parameters to ensure
that we are using the SUDO units effectively.  We also allow the exact
same amount of tuning to the baseline models for fairness.

\subsection {Simple Binary Classification: Checkerboard}
\label {checkerboard}

\begin{figure}[!b]
  \center
   \includegraphics[width=3in]{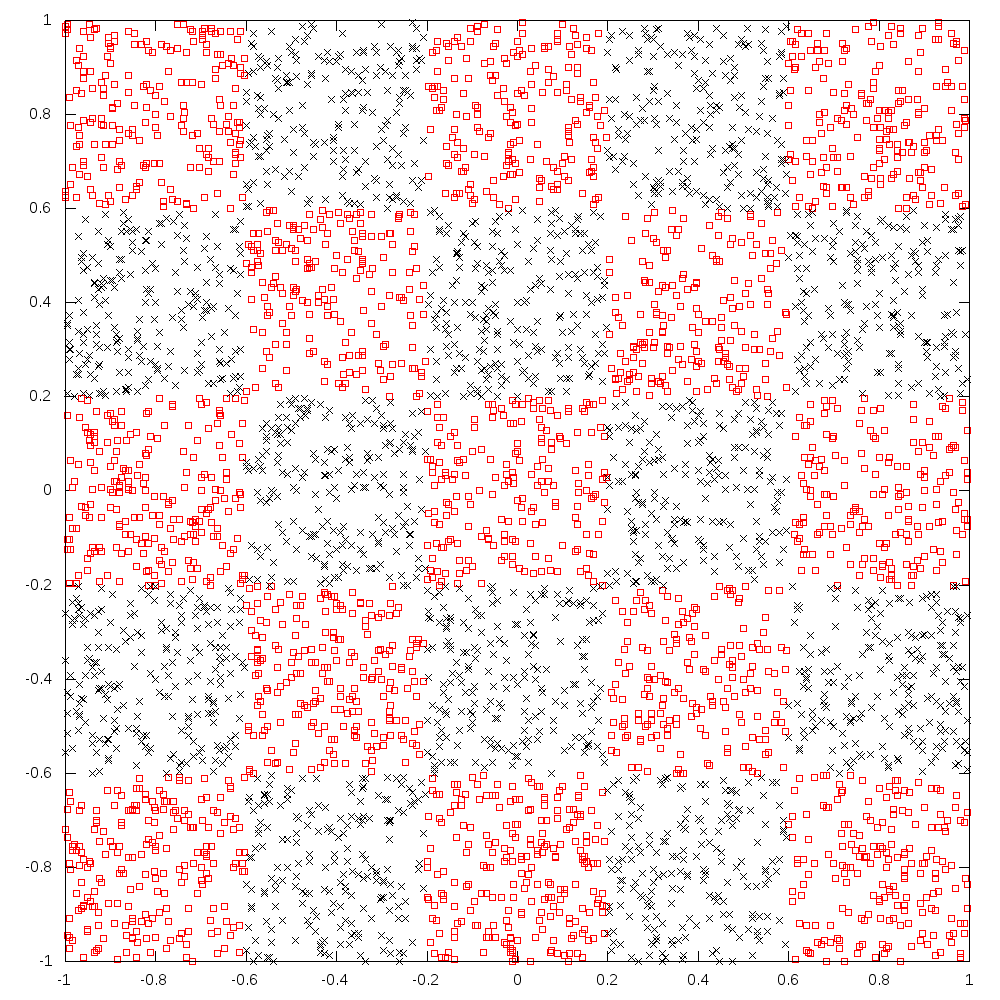}    

  \caption {The 5000 samples chosen for training the binary classification
    checkerboard problem.  This is a real-valued version of the
    classic X-OR problem for training neural networks.}
  \label {fig:checkerboard}
  
\end{figure}

The goal of this problem is to correctly classify points on plane,
based on their real-valued $x,y$ coordinates, as belonging to either a
'black' or 'red' class.  The target classification follows a
checkerboard pattern, as shown in Figure~\ref{fig:checkerboard}.
This is a real-valued version of the traditional X-OR-based problem
used to analyze and study early neural network training~\cite{blum1989training,hertz1991introduction}.

Because we did not know whether the SUDO units would work well in a
single layer or multiple, or even how many units should be used per
layer, a very large variety of experiments are performed.  Three basic
architectures are used.  The first has 2 input units, a single hidden
layer with $H$ units of type $A$ and a single, tanh, output unit.  We
tested $H \in \{5,10,20,50,100\} \times A \in
\{tanh,relu,SUDO\{2,4,8,16,32,64,128,256\}\}$.  The results are shown in
Table~\ref{checkerboard1}.  Because this is a new training
regime, three learning rates (0.001, 0.0001, 0.00001) were attempted
with each activation/architecture combination and the best learning
rate selected for each experiment. Each experiment is replicated three times.  The results in the table show the average performance for the best setting, chosen individually for each experiment.  Note that in this problem, as well as all the others explored in this paper, the relu and tanh units were given the exact same parameter tuning setup.   The accuracies are measured on a set of 250,000 uniformly spaced points.

\begin{table}
  ~
  \center
\caption {Checkerboard Accuracies with 1 Hidden Layer}
\begin{tabular}{r||c|c|c|c|c}
  \toprule
  & \multicolumn{5}{c}{Hidden Units Per Layer}\\
  Activation & 5 & 10 & 50 &100 & 200\\
  \midrule
  \midrule  
 
tanh
&
57.5\%&
73.7\%&
59.7\%&
53.4\%&
54.1\%\\
relu
&
56.5\%&
70.4\%&
85.3\%&
90.6\%&
87.7\%\\
\midrule
sudo-2
&
56.7\%&
69.4\%&
53.1\%&
54.3\%&
53.9\%\\
sudo-4
&
52.7\%&
54.5\%&
55.3\%&
61.1\%&
65.2\%\\
sudo-8
&
58.9\%&
69.1\%&
84.3\%&
80.5\%&
69.0\%\\
sudo-16
&
60.8\%&
73.4\%&
73.1\%&
91.7\%&
88.4\%\\
sudo-32
&
57.2\%&
70.9\%&
84.4\%&
64.9\%&
60.6\%\\
sudo-64
&
60.4\%&
68.2\%&
73.1\%&
57.5\%&
53.3\%\\
sudo-128
&
61.1\%&
73.7\%&
56.8\%&
63.1\%&
55.0\%\\
sudo-256
&
65.0\%&
78.7\%&
68.7\%&
55.8\%&
54.8\%\\

\end{tabular}
\label{checkerboard1}
\end {table}

\begin{figure}
\center
\begin{subfigure}{0.301\textwidth}
   \includegraphics[width=0.999in]{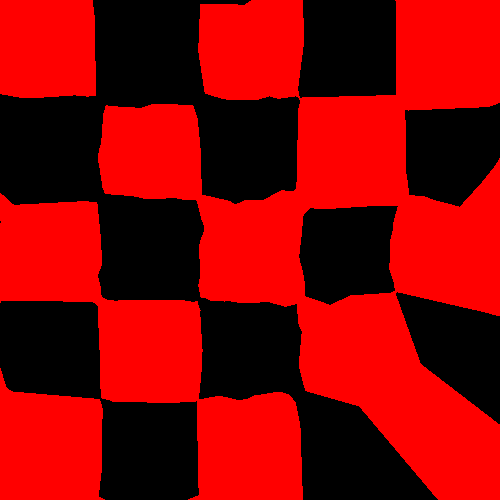}
   \caption{H.Layers=1, 200 Units\\relu (0.93\%)}
\end{subfigure}
\begin{subfigure}{0.301\textwidth}
   \includegraphics[width=0.999in]{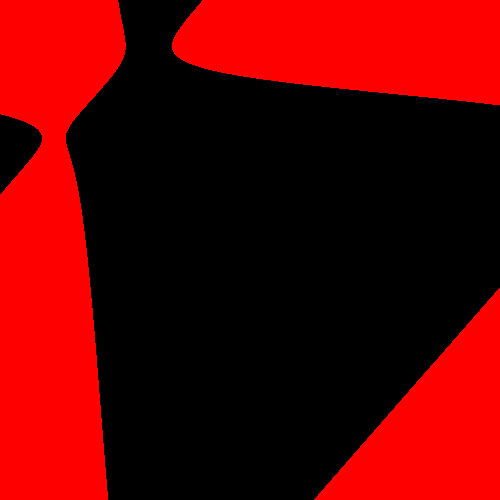}
   \caption{H.Layers=1, 200 Units\\tanh (0.56\%)}
\end{subfigure}
\begin{subfigure}{0.301\textwidth}
   \includegraphics[width=0.999in]{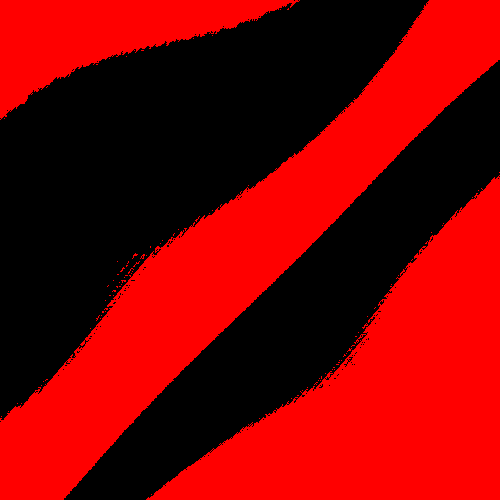}
   \caption{H.Layers=1, 200 Units\\sudo-256 (0.64\%)}
\end{subfigure}
\\ 
\hrule
~\\
\begin{subfigure}{0.301\textwidth}
   \includegraphics[width=0.999in]{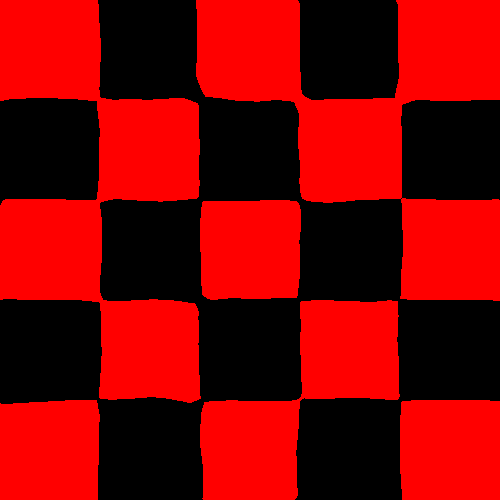}
   \caption{H.Layers=2, 200 Units\\relu (0.98\%)}
\end{subfigure}
\begin{subfigure}{0.301\textwidth}
   \includegraphics[width=0.999in]{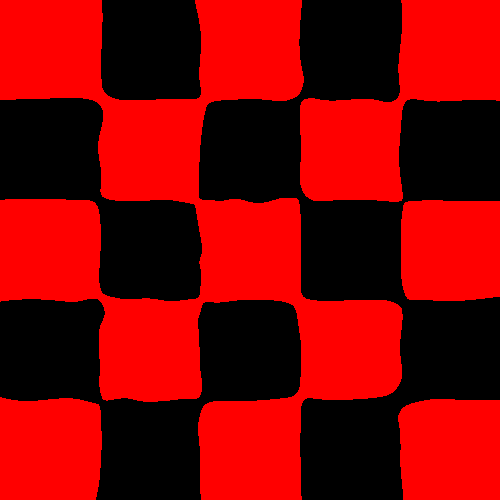}
   \caption{H.Layers=2, 200 Units\\tanh (0.98\%)}
\end{subfigure}
\begin{subfigure}{0.301\textwidth}
   \includegraphics[width=0.999in]{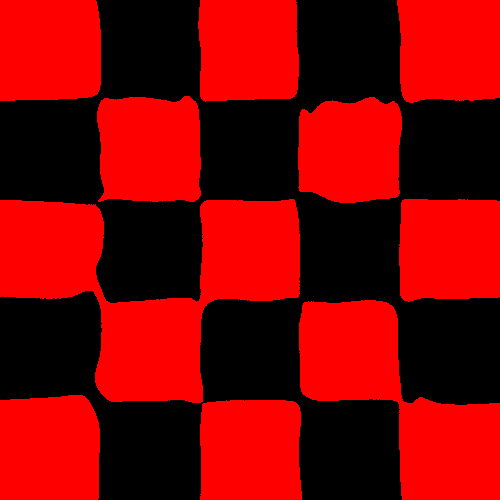}
   \caption{H.Layers=2, 200 Units\\sudo-256 (0.98\%)}
\end{subfigure}
\hrule
~\\
\begin{subfigure}{0.301\textwidth}
   \includegraphics[width=0.999in]{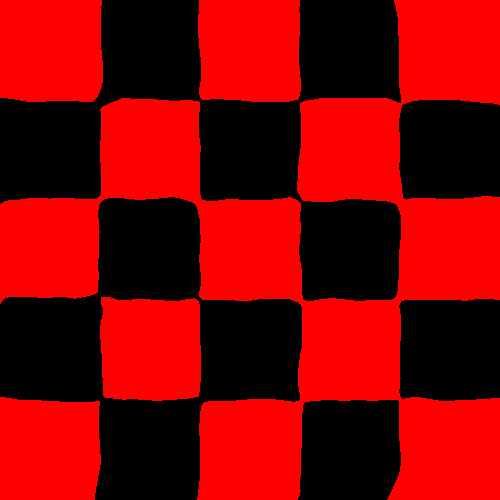}
   \caption{H.Layers=4, 200 Units\\relu (0.98\%)}
\end{subfigure}
\begin{subfigure}{0.301\textwidth}
   \includegraphics[width=0.999in]{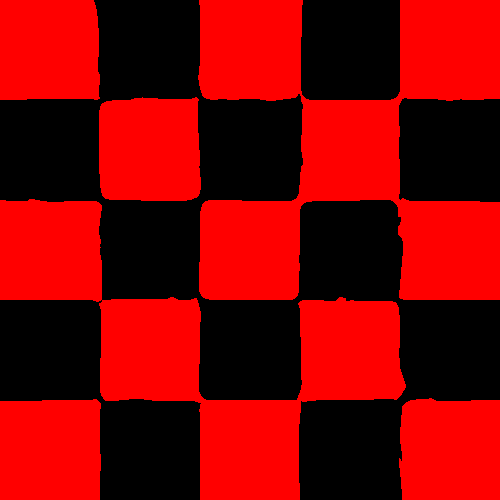}
   \caption{H.Layers=4, 200 Units\\tanh (0.99\%)}
\end{subfigure}
\begin{subfigure}{0.301\textwidth}
   \includegraphics[width=0.999in]{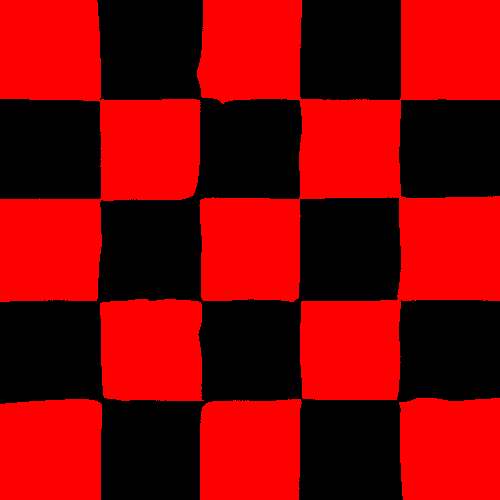}
   \caption{H.Layers=4, 200 Units\\sudo-256 (0.99\%)}
\end{subfigure}
~\\ 
\begin{subfigure}{0.301\textwidth}
   \includegraphics[width=0.999in]{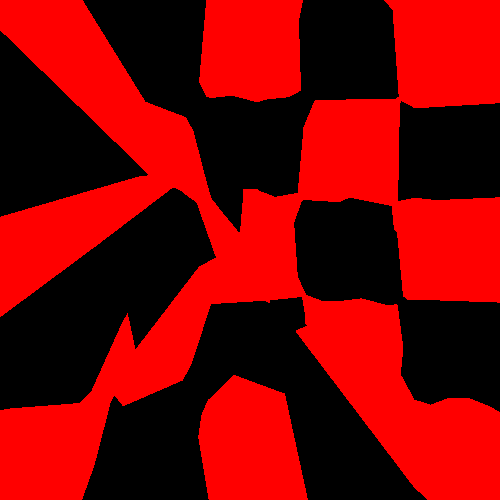}
   \caption{H.Layers=4, 5 Units\\relu (0.88\%)}
\end{subfigure}
\begin{subfigure}{0.301\textwidth}
   \includegraphics[width=0.999in]{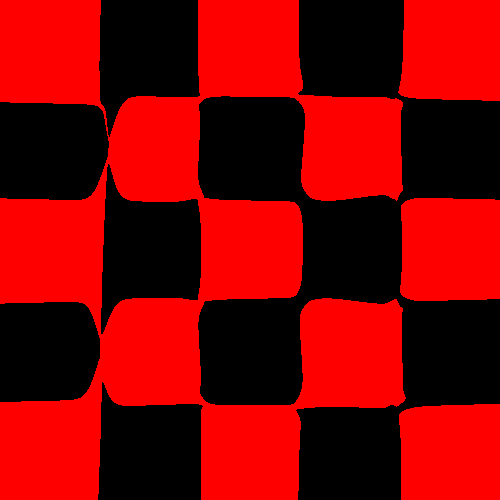}
   \caption{H.Layers=4, 5 Units\\tanh (0.96\%)}
\end{subfigure}
\begin{subfigure}{0.301\textwidth}
   \includegraphics[width=0.999in]{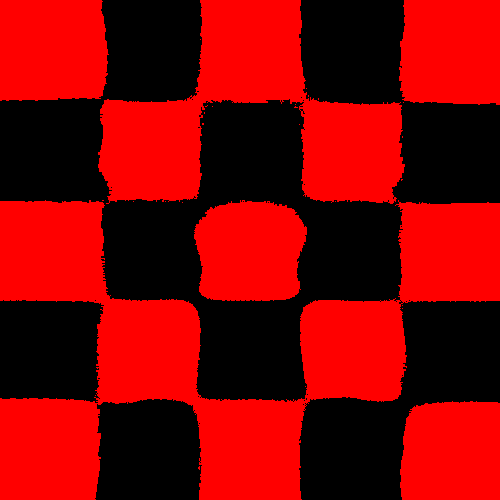}
   \caption{H.Layers=4, 5 Units\\sudo-256 (0.97\%)}
\end{subfigure}

\caption{Decision surfaces for the Checkerboard problem.  (a-c):
networks with a single hidden layer with 200 units.  Relu, tanh, and
SUDO-256 activation units shown.  (g-f) Networks with 2
hidden layers.  (g-l) Networks with 4 hidden layers, with 200 and 5
hidden units per layer. }
\vspace{0.1in}
\label{fig:checkerboard1}
\end{figure}

With a single hidden layer, the relu network performs the best when
given a large number of hidden units.  For the majority of settings of
$L$, the SUDO units perform more similarly to the tanh units;
\textbf{this is a general trend that will be observed in most of the problems
  explored}.  Training with the larger networks did not provide a
substantial benefit to either tanh or SUDO units.  Next, lets examine
what happens when the number of hidden layers is
increased.\footnote{An interesting note to this problem is that with
  a single hidden layer, SUDO-8/16/32 performed well while other $L$
  settings did not.  This is likely due to the number of decision
  boundaries that need to be placed to solve this problem accurately
  (see Figure~\ref{fig:checkerboard1}(d,e)). We did not find this to
  be a trend in other problems.}

In the second architecture, a similar network was created to the
first, but with 2 identically sized hidden layers.  The same variants
of $H$ and $A$ were explored for this architecture.  All the layers are fully connected to the previous layer,
with no skip connections between layers. The results are shown
in Table~\ref{checkerboard2}.  Here, the results have dramatically
changed.  First, note that tanh and SUDO-256 match or outperform relu
units.  Further, despite the fact that SUDO-256 can only output 256
unique values, tanh does \emph{not} perform better.  In fact, reducing
$L$ to between 32 and 64 often rivals the best performances.

What about simple binary outputs or other lower cardinality units?
Looking at SUDO-2 (and SUDO-4), we see that although they do not match
the performance of the other activations when given a small number of
hidden units per layer, as the number of hidden units increases, the
simplest binary activations also perform well.

\begin{table}[t]
  ~
  \center
  \caption {Checkerboard Accuracies with 2 Hidden Layers}
\begin{tabular}{r||c|c|c|c|c}
  \toprule
  & \multicolumn{5}{c}{Hidden Units Per Layer}\\
  Activation & 5 & 10 & 50 &100 & 200\\
  \midrule
  \midrule  
 
tanh
&
81.3\%&
97.5\%&
98.1\%&
97.8\%&
97.0\%\\
relu
&
71.9\%&
91.1\%&
98.0\%&
97.7\%&
97.8\%\\
\midrule
sudo-2
&
54.4\%&
71.1\%&
97.4\%&
98.4\%&
97.6\%\\
sudo-4
&
70.8\%&
95.8\%&
98.7\%&
97.0\%&
96.6\%\\
sudo-8
&
77.8\%&
94.9\%&
97.8\%&
96.9\%&
97.0\%\\
sudo-16
&
81.8\%&
96.4\%&
97.6\%&
97.6\%&
96.8\%\\
sudo-32
&
81.0\%&
96.9\%&
98.0\%&
97.7\%&
97.0\%\\
sudo-64
&
81.7\%&
97.7\%&
98.0\%&
98.0\%&
97.1\%\\
sudo-128
&
83.2\%&
97.2\%&
98.2\%&
97.6\%&
97.5\%\\
sudo-256
&
84.1\%&
96.8\%&
98.0\%&
97.9\%&
96.7\%\\

\end{tabular}
\label{checkerboard2}
\end {table}

In the third architecture, we examine the performance of the SUDO
units in a deeper network.  Here, the network has 4 hidden layers,
each with the same number of hidden units as in previous experiments.
The results are consistent with those found earlier: even a low number
of discrete levels, $L \ge 64$, matches the best performance.  The SUDO
units are able to keep up, and even surpass, the full resolution tanh
and relu units.

\begin{table}[h]
  ~
  \center
  \caption {Checkerboard Accuracies with 4 Hidden Layers}
\begin{tabular}{r||c|c|c|c|c}
  \toprule
  & \multicolumn{5}{c}{Hidden Units Per Layer}\\
  Activation & 5 & 10 & 50 &100 & 200\\
  \midrule
  \midrule  
 
tanh
&
92.8\%&
98.1\%&
98.4\%&
98.4\%&
98.3\%\\
relu
&
81.8\%&
97.1\%&
97.9\%&
97.8\%&
98.1\%\\
\midrule
sudo-2
&
53.8\%&
67.9\%&
89.8\%&
96.1\%&
96.8\%\\
sudo-4
&
66.3\%&
91.5\%&
98.4\%&
97.2\%&
97.3\%\\
sudo-8
&
80.7\%&
97.4\%&
98.2\%&
98.0\%&
98.0\%\\
sudo-16
&
88.4\%&
98.0\%&
98.3\%&
98.1\%&
98.0\%\\
sudo-32
&
91.1\%&
98.1\%&
98.2\%&
97.7\%&
97.8\%\\
sudo-64
&
92.1\%&
98.5\%&
98.4\%&
98.4\%&
97.9\%\\
sudo-128
&
91.0\%&
97.7\%&
98.3\%&
98.4\%&
98.3\%\\
sudo-256
&
93.5\%&
98.3\%&
98.4\%&
98.4\%&
98.3\%\\

\end{tabular}
\label{checkerboard4}
\end {table}

Beyond the quantitative error measurements, it is illuminating to
examine the decision surfaces created by the trained networks, see
Figure~\ref{fig:checkerboard1}.  With a single hidden layer, the relu
layers perform better than the tanh or SUDO units; samples
with a single hidden layer are shown in the first row of
Figure~\ref{fig:checkerboard1}.  

The second row shows that when multiple hidden layers are employed,
the SUDO units (across all of the $L$ discretization levels above 32)
perform similarly to relu and tanh.

In the final two row of Figure~\ref{fig:checkerboard1}, results with 4
hidden layers are shown.  With 200 units, the results look very close
to perfect.  In the last row, with only 5 hidden units, relu appears
significantly worse than with 200 hidden units and significantly worse
than SUDO-256.

\begin{figure}
\center
\begin{subfigure}{0.301\textwidth}
   \includegraphics[width=1in]{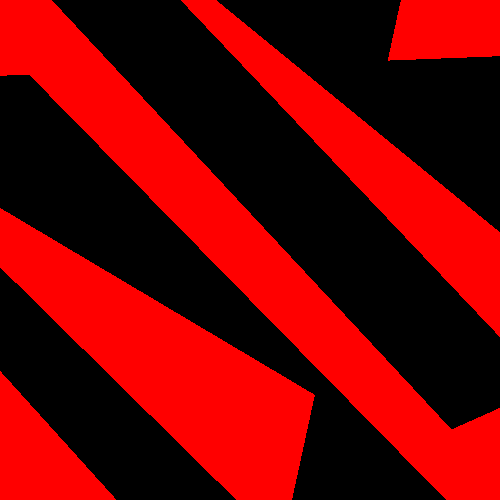}
   \caption{H.Layers=4, 10 Units\\sudo-2 (0.75\%)}
\end{subfigure}
\begin{subfigure}{0.301\textwidth}
   \includegraphics[width=1in]{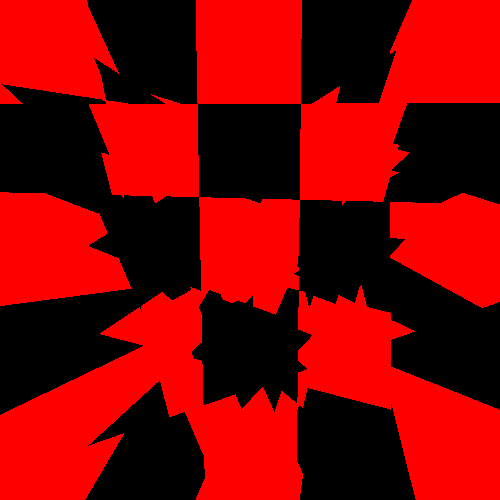}
   \caption{H.Layers=4, 50 Units\\sudo-2 (0.91\%)}
\end{subfigure}
\begin{subfigure}{0.301\textwidth}
   \includegraphics[width=1in]{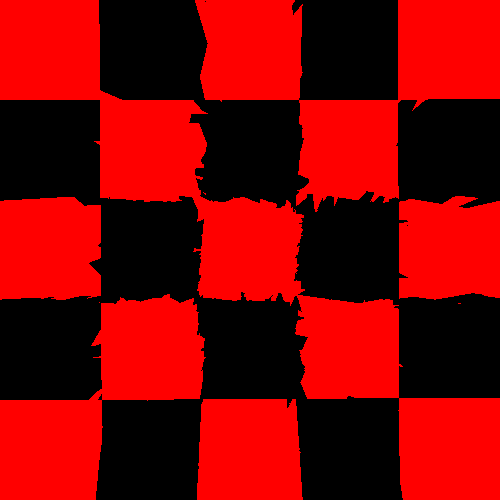}
   \caption{H.Layers=4, 200 Units\\sudo-2 (0.97\%)}
\end{subfigure}
\vspace{-0.1in}
\caption{Decision surfaces for networks with 4 hidden layers and binary activations in the hidden units (SUDO-2).  Shown (a-c) with varying numbers of hidden units 10,50 and 200.}

\label {fig:checkerboard2}

\end{figure}

In the above description, we have largely concentrated on the SUDO
units with $L=256$.  How do SUDO units with $L=2$ carve the decision
surface?  In Figure~\ref{fig:checkerboard2}, results with 10, 50 and
200 units are shown.  With 200 units, even the binary units perform
well.  In contrast, with only 10 units, the boundaries
are less aligned with the underlying distribution.

\clearpage
\subsection {Simple Regression}

An important unanswered question remains after examining only 
classification problems, such as the Checkerboard problem described in
the previous section.  Does the discretized nature of the activation
provide an advantage over continuous activations by easily creating
sharp decision boundaries? And, if that advantage is present, then will the
SUDO units fare worse when asked to \emph{smoothly} approximate response
surfaces?  In this experiment,  we examine how well the SUDO units can be used to
approximate smooth curves.  The goal is to understand whether it is
possible for the networks with discrete outputs to represent a
regression function as accurately as is represented with continuous
and relu units.

\begin{figure}[!b]
\center
  \includegraphics[width=3in]{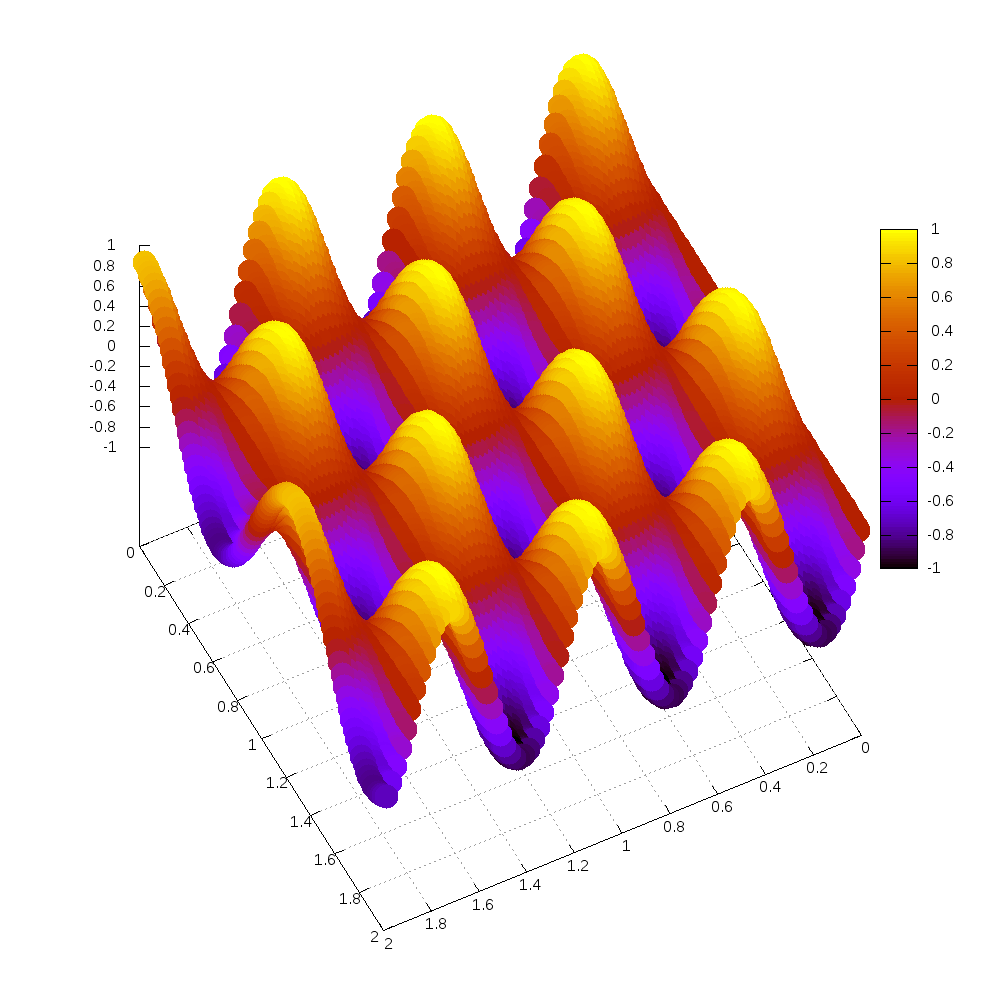}  
  \caption {The surface to be approximated: $z = sin(x*10.0) * cos
    (y*5.0)$.}
  \label {fig:sincos}  
\end{figure}

Consider the function $z = sin (x * 10.0) * cos (y * 5.0)$, in the
range of $x=[-1,1]$ and $y=[-1,1]$.  (See Figure~\ref{fig:sincos}).
How does the discretized output affect the ability of the network to
approximating this surface?

Two basic network architectures were employed for these experiments.
The first had 2 hidden layers and the second had 4.  For each
architecture, the number of hidden units ($H$) in each each hidden
layer was varied $H \in \{10,20,50\}$.  Three learning rates were
attempted for each experiment (0.001, 0.0001, 0.00001).  The best
learning rate was used for each setting (\emph{e.g.} each cell of the
table).  Each trial was replicated 5 times with random weights.  The results are shown in
Tables~\ref{approximate2} \&~\ref{approximate4}.

\begin{table}
  ~
  \center
\caption {Error in approximating a real-value function.  Networks have 2
  hidden layers. }
\begin{tabular}{r||c|c|c}
  \toprule
  & \multicolumn{3}{c}{Hidden Units Per Layer}\\
  Activation & 10 & 20 & 50 \\
  \midrule
  \midrule

tanh
&
0.54&
0.23&
0.11\\
relu
&
0.92&
0.36&
0.11\\
\midrule
sudo-2
&
13.72&
7.70&
3.39\\
sudo-4
&
6.69&
2.04&
0.78\\
sudo-8
&
3.67&
0.96&
0.38\\
sudo-16
&
2.51&
0.60&
0.21\\
sudo-32
&
1.40&
0.48&
0.13\\
sudo-64
&
1.12&
0.37&
0.10\\
sudo-128
&
0.85&
0.32&
0.14\\
sudo-256
&
0.71&
0.25&
0.12\\

\end{tabular}
\label{approximate2}
\end {table}

Compared with the previous experiments, here, the setting of $L$ has
an exaggerated role in the performance of the approximation ---
especially with networks with 10 and 20 hidden units.  Small values of
$L$ perform significantly worse than larger $L$ and tanh and relu
units.  Intuitively, this makes sense as with fewer hidden units,
there are fewer ways of combining the limited number of output values
to approximate the real values.

Another important method of looking at these results is 
where the network's memory ``bits'' are allocated.  For example, if we
allow 160 bits of state to be kept, how should they be
allocated, given Table~\ref{approximate2}?  One possibility is to
create a network with 2 hidden layers, 20 hidden units per layer and
$L=16$.  From the table, we see that this would yield an error of
0.60.  Alternatively, if we use only 10 hidden units in the same
architecture, we could double the bits per hidden unit.  However, that would yield a larger error.  The number of bits
allocated for the representation is \emph{not} sufficient to predict
the performance.  How the bits are distributed (the architecture and discretization) must be considered.

In Table~\ref{approximate4}, the experiment is repeated with 4 hidden
layers.  Again, for each cell, the best learning rate is determined
independently.  Beyond the improved performance across all of the
experiments, notice that the SUDO units ($L=256$) are able to perform
similarly to relu and tanh units.  As before, the more bits that we
allocate to increasing $L$ (while keeping everything else constant)
improves performance.  Even with 50 units, note that the performance
of simple binary units (SUDO-2), as well as other low $L$ settings, remains relatively poor. 

\begin{table}
  ~  
  \center  
\caption {Error in approximating a real-value function.  Networks have 4
  hidden layers. }
\begin{tabular}{r||c|c|c}
  \toprule
  & \multicolumn{3}{c}{Hidden Units Per Layer}\\
  Activation & 10 & 20 & 50 \\
  \midrule
  \midrule

tanh
&
0.12&
0.02&
0.01\\
relu
&
0.33&
0.09&
0.02\\
\midrule
sudo-2
&
15.84&
7.85&
4.09\\
sudo-4
&
8.20&
2.39&
0.69\\
sudo-8
&
2.77&
0.82&
0.32\\
sudo-16
&
1.20&
0.41&
0.16\\
sudo-32
&
0.87&
0.30&
0.08\\
sudo-64
&
0.35&
0.16&
0.08\\
sudo-128
&
0.21&
0.07&
0.04\\
sudo-256
&
0.19&
0.04&
0.02\\

\end{tabular}
\label{approximate4}
\end {table}

\clearpage
\subsection {Effective Network Capacity / Memorization}
\label{memorization}

Though the use of networks as simple associative memory
devices~\cite{carpenter1989neural,specht1988probabilistic,hornik1989multilayer,palm2013neural}
has largely fallen out of research favor, comparing how much a network
can memorize with different activation functions may yield insight
into the relative size of networks needed.  Here, we examine how
accurately a network can reconstruct an image given only each pixel's
$[x,y]$ coordinate.  Unlike the other tasks presented here, as well as
those most commonly explored in the research literature, there is
no explicit notion of generalization.  The network is trained on a single grayscale image and then queried with an arbitrary pixel's coordinates to retrieve its intensity.~\footnote{Note that this task is not the same as
  autoencoding images (autoencoding will be explored in
  Section~\ref{autoencoding}). For autoencoding, the entire image is
  presented at once, and usually many images can be reconstructed by
  the network.  In contrast, for these experiments, the network is
  only given the coordinates of the pixel and must recall the pixel's
  correct intensity value for the single training image.}

\begin {figure}

\center
  \begin{subfigure}{\textwidth}
    \center    
    \includegraphics[width=2.5in]{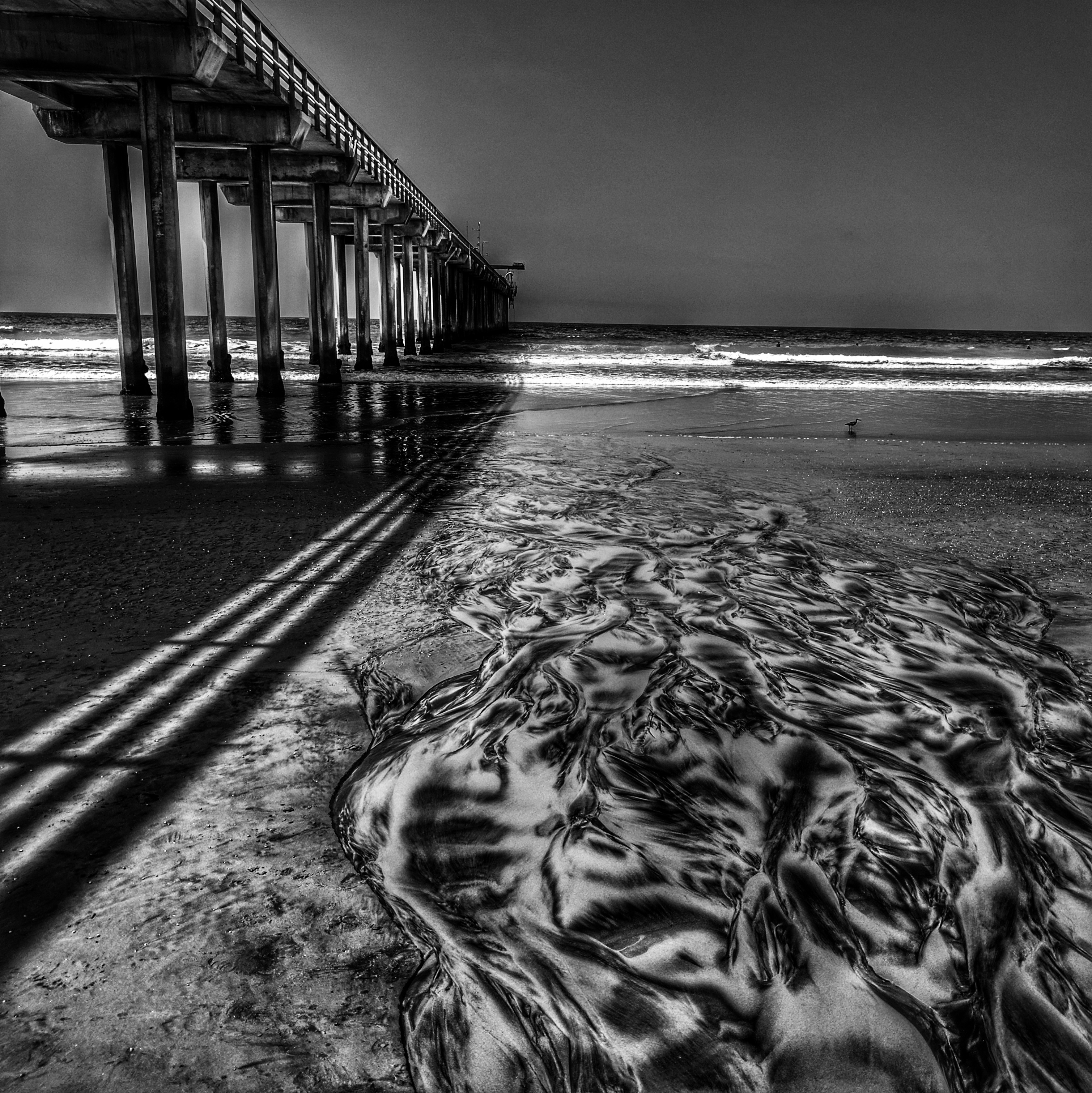} 
    \caption{original}
  \end{subfigure}\\
~\\
  \begin{subfigure}{0.24\textwidth}
    \center
    \includegraphics[width=1.3in]{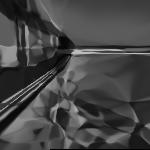}
    \caption{tanh}
  \end{subfigure}
  \begin{subfigure}{0.24\textwidth}
    \center
    \includegraphics[width=1.3in]{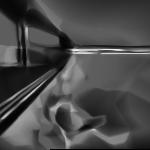}
    \caption{relu}
  \end{subfigure}
  \\
~\\
  \begin{subfigure}{0.24\textwidth}
    \center
\includegraphics[width=1.3in]{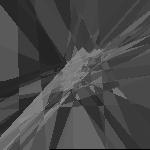}    
    \caption{SUDO-2}
  \end{subfigure}
  \begin{subfigure}{0.24\textwidth}
      \center
\includegraphics[width=1.3in]{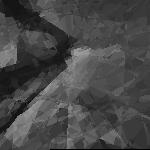}      
    \caption{SUDO-4}
  \end{subfigure}
  \begin{subfigure}{0.24\textwidth}
        \center
\includegraphics[width=1.3in]{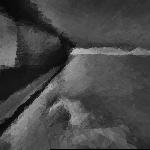}        
    \caption{SUDO-8}
  \end{subfigure}
  \begin{subfigure}{0.24\textwidth}
          \center
\includegraphics[width=1.3in]{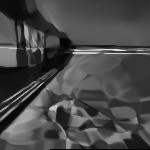}          
    \caption{SUDO-256}
  \end{subfigure}

  \caption {Performance on the memorization task.  Original image of pier and water, composed of
    $150\times150$ pixels, shown on top.  Networks have 4 hidden
    layers.  SUDO-2 (d) yields an unrecognizable image.  With 4
    discretization levels (e) the dock becomes recognizable.  By 256,
    the reconstruction has vastly improved.}
  \label{fig:memresults}
\end{figure}

In these experiments, the original image is a grayscale intensity
image of size $150\times150$. The image has been post-processed with
pseudo-HDR to ensure variation in intensity values.  It is shown in
Figure~\ref{fig:memresults}(a).

The results are \underline{\textbf{not}} indicative of a theoretical measure of network
capacity.  They are, however, indicative of the practical accessible
capacity given a particular learning approach.  The learning approach
(in this case ADAM~\cite{kingma2014adam} and/or SGD+Momentum) has a
very large role in determining how the network represents the
information -- both the form and efficiency.  Different learning
approaches may be able to better utilize a network's
capacity. Nonetheless, we propose that this is an interesting
experiment as (1) the tested optimization procedures are the most
commonly used.  And, (2), the same learning procedure is used across
all of the trials with all of the hidden units, yielding relative
results valid for comparison.

\begin{table}[!h]
  ~
  \center
\caption {SSE memorizing image.  Networks have 1
  hidden layer. Last Activation is tanh.}
\begin{tabular}{r||c|c|c}
  \toprule
  & \multicolumn{3}{c}{Hidden Units Per Layer}\\
  Activation & 50 & 100 & 200 \\
  \midrule
  \midrule

tanh
&
1509.18&
1535.56&
1547.50\\
relu
&
1436.83&
1386.76&
1370.26\\
\midrule
sudo-2
&
1702.52&
1715.54&
1750.73\\
sudo-4
&
1648.93&
1644.20&
1649.05\\
sudo-8
&
1608.58&
1634.98&
1597.16\\
sudo-16
&
1541.39&
1535.25&
1533.82\\
sudo-32
&
1498.50&
1511.71&
1546.28\\
sudo-64
&
1529.45&
1530.31&
1563.23\\
sudo-128
&
1520.51&
1541.00&
1544.91\\
sudo-256
&
1530.65&
1529.95&
1568.84\\

\end{tabular}
\label{memorize1}
\end {table}

When using a single hidden layer, the discrete activations can achieve
similar performance to the tanh units across all sizes of hidden
layers. However, discretization levels between 32-256 are required.
The relu units have a clear advantage over both.  Next, in Table~\ref{memorize2}
and Table~\ref{memorize4}, we examine the performance with 2 and 4
hidden layers.  One of the peculiarities of these results is that for
tanh, a larger network did not consistently yield improved results.
Since the performance of SUDO units often approximates tanh, this
trend also extended to many SUDO settings.  This is likely due to the
settings of the training hyper-parameters.  Nonetheless, we did not
change them for these experiments to keep them consistent across the
paper.

\begin{table}[!h]
  ~
  \center
\caption {SSE memorizing image.  Networks have 2
  hidden layers. Last Activation is tanh.}
\begin{tabular}{r||c|c|c}
  \toprule
  & \multicolumn{3}{c}{Hidden Units Per Layer}\\
  Activation & 50 & 100 & 200  \\
  \midrule
  \midrule

tanh
&
1285.85&
1289.60&
1319.10\\
relu
&
1201.02&
1152.29&
1145.68\\
\midrule
sudo-2
&
1505.62&
1454.44&
1370.68\\
sudo-4
&
1446.20&
1448.04&
1462.08\\
sudo-8
&
1367.80&
1424.81&
1458.70\\
sudo-16
&
1265.13&
1323.53&
1350.90\\
sudo-32
&
1266.70&
1326.84&
1355.29\\
sudo-64
&
1267.70&
1292.24&
1372.25\\
sudo-128
&
1229.46&
1303.49&
1349.90\\
sudo-256
&
1233.91&
1298.23&
1301.24\\

\end{tabular}
\label{memorize2}
\end {table}

\begin{table}[]
  ~
  \center
\caption {SSE memorizing image.  Networks have 4 hidden layers. Last Activation is tanh.}
\begin{tabular}{r||c|c|c}
  \toprule
  & \multicolumn{3}{c}{Hidden Units Per Layer}\\
  Activation & 50 & 100 & 200   \\
  \midrule
  \midrule

tanh
&
878.56&
895.68&
935.90\\
relu
&
937.17&
871.23&
852.40\\
\midrule
sudo-2
&
1479.91&
1418.88&
1372.66\\
sudo-4
&
1292.34&
1263.84&
1334.85\\
sudo-8
&
994.39&
1168.80&
1244.86\\
sudo-16
&
916.12&
988.33&
1147.57\\
sudo-32
&
917.15&
923.70&
1071.47\\
sudo-64
&
889.95&
911.23&
996.22\\
sudo-128
&
888.76&
872.15&
986.37\\
sudo-256
&
918.70&
884.72&
964.91\\

\end{tabular}
\label{memorize4}
\end {table}

In Tables~\ref{memorize2} \&~\ref{memorize4}, the performances of tanh
and SUDO-256 are similar: relu continues to outperform both.  This is
the largest difference in performance between tanh/SUDO and relu
witnessed in this paper.  To see if the disparity continues, two extra
experiments are constructed.  First, does the advantage continue with
deeper architectures?  Table~\ref{memorize10} shows the results with a
network of depth 10.  We see that relu units again outperform
both tanh and SUDO results.  The increased depth does not equalize the
performance of the tanh or the SUDO units with the relu
activations. How do we regain the lost performance in comparison to
relu?  We can ``rectify'' the SUDO units as well; see the appendix
for in-depth details and experiments.

\begin{table}
  ~
  \center
\caption {SSE memorizing image.  Networks have 10
  hidden layers. Last Activation is tanh.}
\begin{tabular}{r||c|c|c}
  \toprule
  & \multicolumn{3}{c}{Hidden Units Per Layer}\\
  Activation & 50 & 100 & 200   \\
  \midrule
  \midrule

tanh
&
885.43&
939.55&
857.40\\
relu
&
769.27&
745.33&
662.66\\
\midrule
sudo-2
&
1625.15&
1573.95&
1545.70\\
sudo-4
&
1412.62&
1373.34&
1398.23\\
sudo-8
&
1165.48&
1116.34&
1107.36\\
sudo-16
&
1024.29&
965.74&
968.67\\
sudo-32
&
927.06&
933.10&
930.50\\
sudo-64
&
929.23&
856.97&
881.00\\
sudo-128
&
909.19&
892.16&
887.31\\
sudo-256
&
903.90&
889.50&
874.53\\

\end{tabular}
\label{memorize10}
\end {table}

\begin {figure}

\center
  \begin{subfigure}{0.24\textwidth}
    \center    
    \includegraphics[width=1.5in]{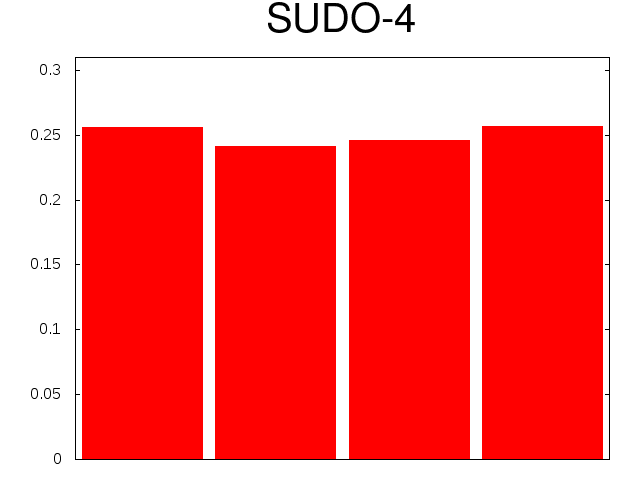}
    \caption{}
  \end{subfigure}
  \begin{subfigure}{0.24\textwidth}
    \center    
    \includegraphics[width=1.5in]{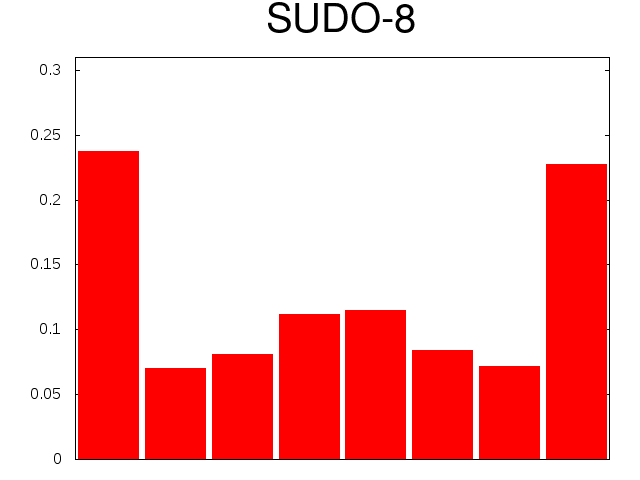}
    \caption{}    
  \end{subfigure}
  \begin{subfigure}{0.24\textwidth}
    \center    
    \includegraphics[width=1.5in]{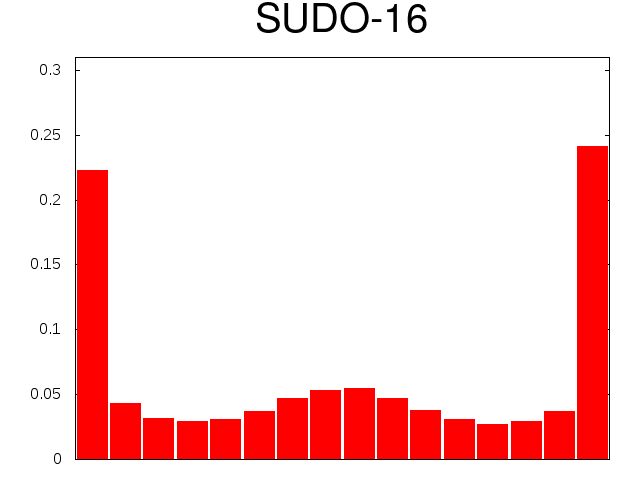}
    \caption{}        
  \end{subfigure}
  \begin{subfigure}{0.24\textwidth}
    \center    
    \includegraphics[width=1.5in]{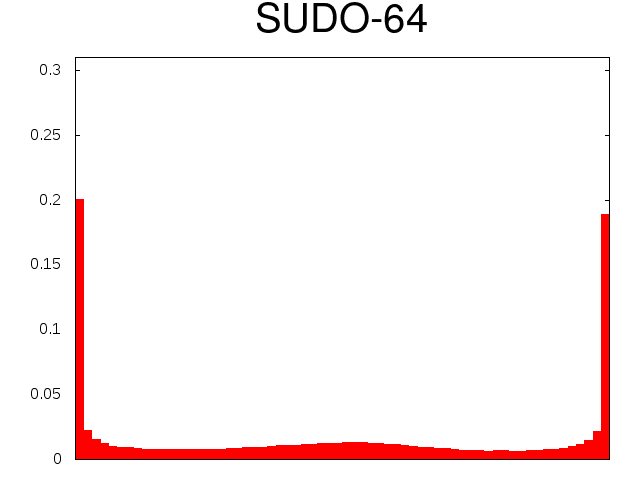}
    \caption{}            
  \end{subfigure}\\
  ~\\
  ~\\    
  ~\\  
  \begin{subfigure}{0.24\textwidth}
    \center    
    \includegraphics[width=1.5in]{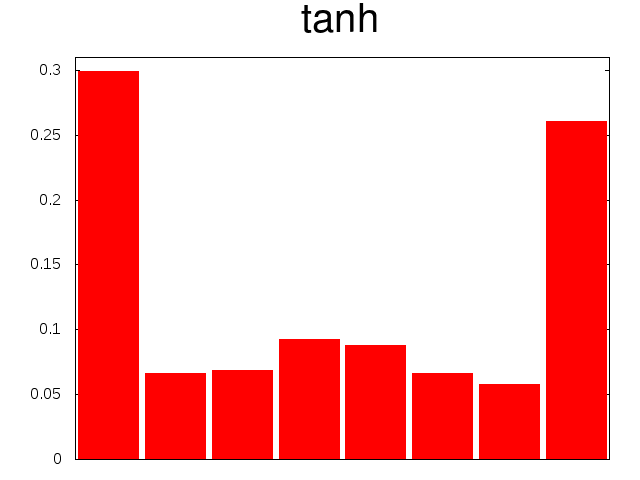}
    \caption{tanh (discretized to 8 levels)}                
  \end{subfigure}\\
  \caption {Distributions of hidden unit output activations,
    post-training.  Histogram shows the results across all layers of
    the 4-hidden layers memorization networks, trained with 50 hidden
    units. SUDO-{4,8,16,64} and tanh (discretized to 8 equal sized
    bins) shown. }
  \label{fig:memdists}
  
\end{figure}

Second, we look at the distribution of activations for the SUDO units
(across all of the SUDO hidden units in the networks
trained with 4 hidden layers and 50 hidden units per layer).  This is
compared to the activation of the tanh units when discretized into 8
equal sized bins.  We see a common trend: in the discretized and
non-discretized units, the extrema get the most activations.
The histograms are shown in Figure~\ref{fig:memdists}.

To see how well each network does in the task of reconstruction,
Figure~\ref{fig:memresults} shows the reconstruction of the original
image achieved by the networks.

\newpage

\clearpage
\subsection {Autoencoding}
\label {autoencoding}

In this section, we examine the standard task of autoencoding images.
For these tests, networks were trained on ImageNet images, scaled to
32x32, and tested on an independent test set, also from
ImageNet~\cite{deng2014imagenet}.

For these experiments, two very different network architectures were
used: one with a series of convolutions and the other fully connected.
As described below, the ``bottleneck'' of the convolution is much
larger in the conv-nets than in the fully connected nets and thereby
the convolution-net outperformed the fully connected network for all
comparable settings.  No attempt was made to balance the two
architecture's bottleneck sizes as, for this study, the sole important
comparison is between the hidden unit activations of the same
architecture.

Table~\ref{autoConv} shows the results with the convolution
architecture.  The first set of experiments were conducted with 4
layers of 3x3 convolutions with stride 2x2.  The layers had 50, 50,
40, and 20 filters per layer.  This was followed by 4 layers in the
decode step of 2d-transpose (deconvolution layers) of depth 40, 50,
50, 20 and finally a 3 (RGB) layer for recreating the image.  The
final activations of the outputs were a tanh.  

\begin{table}[h!]
  \centering  
  \parbox{0.45\linewidth}{
    \caption {Autoencoding Networks Convolution-Based - Relative SSE Errors}
    \vspace{-0.05in}
\begin{tabular}{r||c|c|c|c}
  \toprule
  & \multicolumn{4}{c}{Network Scale}\\
  Activation & 1 & $\times2$ & $\times4$ & $\times8$\\  
  \midrule
  \midrule  
 
tanh
&
1.00&
0.68&
0.35&
0.06\\
relu
&
1.02&
0.69&
0.37&
0.09\\
\midrule
sudo-2
&
1.92&
1.58&
1.16&
0.87\\
sudo-4
&
1.41&
1.01&
0.68&
0.35\\
sudo-8
&
1.13&
0.79&
0.46&
0.17\\
sudo-16
&
1.05&
0.71&
0.38&
0.11\\
sudo-32
&
1.01&
0.68&
0.34&
0.07\\
sudo-64
&
0.99&
0.66&
0.34&
0.07\\
sudo-128
&
0.98&
0.65&
0.33&
0.07\\
sudo-256
&
0.98&
0.65&
0.33&
0.06\\

\end{tabular}
\label{autoConv}
}~~~~~~
\parbox{0.45\linewidth}{
  \caption {Autoencoding Networks, Fully Connected - Relative SSE Errors}
    \vspace{-0.05in}  
\begin{tabular}{r||c|c|c|c}
  \toprule
  & \multicolumn{4}{c}{Network Scale}\\
  Activation & 1 & $\times2$ & $\times4$ & $\times8$\\    
  \midrule
  \midrule  
 
tanh
&
2.29&
1.98&
1.62&
1.30\\
relu
&
2.39&
2.05&
1.81&
1.48\\
\midrule
sudo-2
&
3.51&
3.16&
2.94&
2.67\\
sudo-4
&
2.99&
2.75&
2.40&
2.09\\
sudo-8
&
2.70&
2.36&
2.08&
1.71\\
sudo-16
&
2.53&
2.16&
1.82&
1.49\\
sudo-32
&
2.37&
2.02&
1.70&
1.38\\
sudo-64
&
2.34&
1.99&
1.67&
1.33\\
sudo-128
&
2.32&
1.98&
1.64&
1.31\\
sudo-256
&
2.31&
1.99&
1.64&
1.31\\

\end{tabular}
\label{autoFull}
}
  \end{table}

For ease of reading this table, we set the performance of the conv-net
with a tanh activation function as the baseline.  Accordingly, the
first column of Table~\ref{autoConv} shows the \emph{relative} results
of all of the hidden activation types explored.  The lower the number,
the lower the SSE and therefore the better the performance.  The
subsequent three columns of Table~\ref{autoConv} ($\times2$,
$\times4$, $\times8$) show the performance achieved by increasing the
number of filters per layer (for all layers) by a factor of
$\times2,\times4,\times8$.  As expected, there is a sharp
decrease in the error across all of the hidden unit activations as the
number of filters per layer increases.  Importantly, note that the
good performance, relative to the tanh and relu units is achieved by
SUDO-64, and continues as the discretization levels are increased for
every network size.

We next repeat the same set of experiments using networks with fully
connected layers.  For these experiments, the fully connected
architecture had 4 layers for the encoding; these consisted of 50, 50,
40 and 20 hidden units. Next, 3 layers for decoding had 40, 50, 50
hidden units.  This was followed by a reconstruction layer of size
32x32x3 (RGB) with tanh units.  Table~\ref{autoFull} shows the
results, again relative to the score of the tanh, scale 1, with
conv-nets.  As in Table~\ref{autoConv}, the columns of
Table~\ref{autoFull} ($\times2$, $\times4$, $\times8$) show the
performance relative to the number of hidden units per
layer.  Like
the results with the conv-nets, larger networks have improved
performance.  Also, again, we see little difference in performance
between the SUDO activations and the better of tanh and relu as $L$ is
increased.

\clearpage
\subsection {MNIST Digit Classification}

As a final test, we examine the performance of SUDO units on the
standard MNIST digit classification task~\cite{lecun1998mnist}.  For
this test, we use fully connected networks and vary the number of
hidden units per layer and the number of hidden layers.
Table~\ref{mnist1} shows the results with a network employing a single
hidden layer.  Table~\ref{mnist1} provides a similar set of results
using a network with 4 hidden layers.  Each entry shown in the tables
is the average of 5 networks trained with randomly initialized
starting weights.\footnote{Though alternate
  architectures with convolutions are known to provide state of the
  art results, for clarity with the effects of number of hidden units,
  we use only fully connected layers.  Again, as before, we are most
  interested in seeing the \emph{relative} performance of the SUDO
  units to the tanh and relu activations.}  For this task, very few discretization levels provide
competitive performance when the number of hidden units is $>3$. For
example, SUDO-8 and SUDO-16 often perform as well as tanh and surpass
relu in performance throughout the range of number of hidden units
used in the 4 hidden layer architecture. 

\begin{table}[!h]
  \center
\caption {MNIST Accuracy Results.  Networks have 1 hidden layer. }
\begin{tabular}{r||c|c|c|c|c|c}
  \toprule
  & \multicolumn{6}{c}{Hidden Units Per Layer}\\
  Activation & 2 & 3 &4 & 10 & 50 & 100\\
  \midrule
  \midrule
 
tanh
&
64.7\%&
77.6\%&
85.7\%&
93.6\%&
97.1\%&
97.7\%\\
relu
&
69.7\%&
81.2\%&
86.8\%&
94.1\%&
97.2\%&
97.8\%\\
\midrule
sudo-2
&
38.9\%&
67.8\%&
82.2\%&
90.8\%&
96.1\%&
97.2\%\\
sudo-4
&
56.9\%&
76.6\%&
84.0\%&
92.9\%&
96.8\%&
97.7\%\\
sudo-8
&
63.0\%&
76.9\%&
85.6\%&
93.4\%&
96.9\%&
97.7\%\\
sudo-16
&
61.9\%&
79.6\%&
85.2\%&
93.9\%&
96.9\%&
97.7\%\\
sudo-32
&
64.5\%&
79.4\%&
86.1\%&
93.6\%&
97.1\%&
97.7\%\\
sudo-64
&
64.1\%&
78.6\%&
86.6\%&
93.4\%&
97.1\%&
97.8\%\\
sudo-128
&
61.5\%&
80.7\%&
86.4\%&
93.5\%&
97.3\%&
97.7\%\\
sudo-256
&
63.0\%&
78.1\%&
85.6\%&
93.6\%&
97.2\%&
97.8\%\\

\end{tabular}
\label{mnist1}
\end {table}

\begin{table}[!h]
  \center  
\caption {MNIST Accuracy Results.  Networks have 4 hidden layers. }

\begin{tabular}{r||c|c|c|c|c|c}  
  \toprule
  & \multicolumn{6}{c}{Hidden Units Per Layer}\\
  Activation & 2 &3 &4 & 10& 50& 100\\  
  \midrule
  \midrule
 
tanh
&
55.7\%&
77.4\%&
86.1\%&
94.1\%&
97.2\%&
97.9\%\\
relu
&
33.6\%&
66.7\%&
81.5\%&
93.7\%&
97.2\%&
97.8\%\\
\midrule
sudo-2
&
25.6\%&
36.0\%&
41.4\%&
88.2\%&
95.8\%&
97.2\%\\
sudo-4
&
46.0\%&
66.1\%&
79.9\%&
92.5\%&
96.9\%&
97.8\%\\
sudo-8
&
52.6\%&
71.7\%&
83.0\%&
93.4\%&
97.2\%&
97.9\%\\
sudo-16
&
56.9\%&
70.5\%&
84.1\%&
93.9\%&
97.3\%&
98.0\%\\
sudo-32
&
64.5\%&
74.2\%&
84.1\%&
93.9\%&
97.2\%&
97.8\%\\
sudo-64
&
57.3\%&
75.6\%&
85.4\%&
94.2\%&
97.3\%&
98.0\%\\
sudo-128
&
61.1\%&
75.8\%&
84.3\%&
93.7\%&
97.3\%&
97.9\%\\
sudo-256
&
62.9\%&
75.4\%&
85.0\%&
94.0\%&
97.2\%&
98.1\%\\

\end{tabular}
\label{mnist4}
\end {table}

\clearpage

\section {Discussion \& Future Work}

The most salient finding in this work is that reducing the number of
outputs allowed in a network's hidden units from $2^{32}$ (a typical
floating point representation) to between only 64-256 unique outputs
does not have a noticeable impact on performance.  This is valuable in
many modern scenarios where computation needs to be limited (\emph{e.g.} mobile devices) or where
memory is at a premium, such as pixel recurrent neural networks.

A second interesting trend that was noted is the role of network
depth.  With lower values of $L$, shallow networks exhibited severely
degraded performance.   However, as the depth of the
network was increased (even to modest depth of 2 or 4), networks with
small $L$ exhibited large improvements in performance.   For example,
in the memorization task (Section~\ref{memorization}), doubling the number of units
in a single hidden layer had far less performance improvement compared
to adding another layer with the same number of units.   This is
similar to the performance improvements seen with more common
activations.

Though there has been a significant previous work in ``binarizing''
networks (\emph{e.g.} setting $L=2$), the difficulties that were
experienced in training and using these networks is greatly dissipated
when the discretization level is increased.  With large $L$,  no
modifications to the training algorithms is required.  In this study,
standard TensorFlow with both ADAM and SGD+Momentum were used, with
the exact same settings for training, validation and testing.

In this study, we used the sigmoid or tanh as the underlying
activation function to be discretized.  In the cases in which the SUDO
units did not perform was well as relu, generally, tanh did not
either.  The most straightforward method to address this is to change
the underlying function.  In the appendix, we examine what happens
when the SUDO units are replaced with rectified-SUDO units.
Analogously to relu units, these units emit a 0 for large parts of the
activation.  All of the experiments in this paper are repeated with
rectified-SUDO units.  The performance is much closer to standard relu
units.

The simplicity of the approach opens a number of new potential avenues
for research.  For example, future work should be conducted with units
with different discretization levels ($L$) within the same
network.  As was witnessed when examining the real-valued checkerboard
problems, some of the middle discretization settings performed better
than those both smaller and larger.  A simple method of capturing this
benefit is to use different $L$'s for tackling the same problem,
within the same network.

An alternative approach to \emph{a priori} guessing the right $L$ is
to make this a learnable parameter.  A similar approach was taken
in~\cite{agostinelli2014}, in which a piece-wise linear activation
function was learned for each neuron with promising results when
compared to static rectified linear units.

\bibliography{nips2017}
\bibliographystyle{unsrt}

\begin{appendices}
  \label{appendix}
  \section {Rectified SUDO Units}

  The results reported in this paper have shown that the performance
  of the SUDO units closely tracks the performance of tanh units.
  However, in a few of the problems described in
  Section~\ref{experiments}, the rectified linear units were able to
  outperform tanh (and therefore SUDO).  The most stark case was found
  in the memorization task, Section~\ref{memorization}, as well as a
  few instantiations of the the Checkerboard binary classification
  task, Section~\ref{checkerboard}.

  Here, we briefly describe a simple method to more closely
  approximate the performance of the relu units by ``rectifying'' the SUDO
  units.  The resulting activation of the \emph\textbf{R-SUDO} units
  is as follows:
\vskip 0.25in
\begin{fmpage}{0.9\textwidth}

\begin{algorithmic}
    \STATE function RECTIFIED\_SUDO\_Activation (input, levels):
    \bindent
    \STATE underlying  $\gets$ $tanh (input)$
    \IF {underlying < 0:}
    \STATE return (0);
    \ELSE
    \STATE activation\_step $\gets$ $2 / (levels - 1)$
    \STATE plateauRange    $\gets$ $2 / levels$
    \STATE output $\gets$ $(\left \lceil{ (underlying + 1.0) / plateauRange}\right \rceil- 1.0) * activation\_step$
    \STATE return ($-1.0 + output$)
    \ENDIF
     \eindent    
  \end{algorithmic}
\end{fmpage}
\vskip 0.15in

  Several points should be noted about this simple method of
  rectifying the SUDO units.
  \begin{itemize}

    \item As with relu units, these units propagate no derivatives when $underlying < 0$. 

      \item In this naive implementation, note that the number of
        activations for $b$ bits is reduced from $2^b$ to $(2^{b-1} +
        1)$.  This is wasteful in terms of bits.  Nonetheless, for ease
        of comparison to the rest of the results, and so that the tables in the
        appendix and the main body of the paper can be compared cell
        by cell, we keep this representation.  In practice, to achieve $2^b$ discretization levels, they should
        all be placed in the $\ge 0$ range.

        \item Like the relu-6 variant, this activation has a sharp non-linearity
          and a maximum output activation
          \cite{krizhevsky2010convolutional}.  The use of an
          underlying linear activation, such as with relu-6, can easily be substituted here.

  \end{itemize}

  In the remainder of this section, we recreate all of the experiments
  with R-SUDO units and provide all the results for completeness.  In
  particular, examine the results in Section~\ref{rmemorization}.
  The R-SUDO results now match relu closely.  Additionally, where tanh outperforms relu, SUDO tends to outperform
  R-SUDO.

  When examining the results, note that the experiments with tanh
  and relu are rerun.  Though they should be similar to those
  observed earlier, due to random weight initializations,
  differences may be present.  Any differences give an indication of
  the typical variance experienced in these experiments.  In both the
  main body and in this appendix, each table cell is the average of
  multiple runs, as described in Section~\ref{experiments}.

\clearpage
\subsection {Checkerboard with R-SUDO units}
\begin{table}[h!]
  \center
  \caption {Checkerboard Accuracies with 1 Hidden Layer.  Compare to Table~\ref{checkerboard1}.}
  \footnotesize
\begin{tabular}{r||c|c|c|c|c}
  \toprule
  & \multicolumn{5}{c}{Hidden Units Per Layer}\\
  Activation & 5 & 10 & 50 &100 & 200\\
  \midrule
  \midrule  
 
tanh
&
59.8\%&
80.5\%&
67.7\%&
55.0\%&
53.3\%\\
relu
&
58.0\%&
66.9\%&
86.9\%&
87.9\%&
92.9\%\\
\midrule
r-sudo-2
&
59.1\%&
64.8\%&
80.3\%&
80.2\%&
82.1\%\\
r-sudo-4
&
57.7\%&
66.0\%&
85.8\%&
87.3\%&
89.1\%\\
r-sudo-8
&
61.3\%&
72.1\%&
92.7\%&
93.6\%&
94.3\%\\
r-sudo-16
&
64.5\%&
77.9\%&
94.5\%&
95.8\%&
96.0\%\\
r-sudo-32
&
62.5\%&
77.9\%&
94.7\%&
96.1\%&
96.8\%\\
r-sudo-64
&
66.9\%&
77.7\%&
94.5\%&
96.0\%&
96.9\%\\
r-sudo-128
&
65.2\%&
75.9\%&
94.9\%&
96.3\%&
97.0\%\\
r-sudo-256
&
65.6\%&
76.9\%&
95.3\%&
96.3\%&
96.9\%\\

\end{tabular}
\label{Lcheckerboard1}
\end {table}

\begin{table}[h!]
  \center
  \caption {Checkerboard Accuracies with 2 Hidden Layers. Compare to Table~\ref{checkerboard2}.}
  \footnotesize
\begin{tabular}{r||c|c|c|c|c}
  \toprule
  & \multicolumn{5}{c}{Hidden Units Per Layer}\\
  Activation & 5 & 10 & 50 &100 & 200\\
  \midrule
  \midrule  
 
tanh
&
82.3\%&
97.2\%&
98.2\%&
98.1\%&
97.6\%\\
relu
&
73.9\%&
91.8\%&
97.2\%&
97.9\%&
97.6\%\\
\midrule
r-sudo-2
&
52.9\%&
52.6\%&
54.5\%&
57.9\%&
58.3\%\\
r-sudo-4
&
55.9\%&
75.6\%&
95.6\%&
95.5\%&
94.7\%\\
r-sudo-8
&
65.4\%&
83.2\%&
98.4\%&
97.5\%&
98.4\%\\
r-sudo-16
&
72.4\%&
89.2\%&
98.0\%&
97.6\%&
97.5\%\\
r-sudo-32
&
72.6\%&
91.9\%&
98.0\%&
97.7\%&
97.7\%\\
r-sudo-64
&
76.5\%&
92.7\%&
98.6\%&
98.0\%&
98.0\%\\
r-sudo-128
&
78.0\%&
93.4\%&
98.2\%&
98.0\%&
98.1\%\\
r-sudo-256
&
76.4\%&
94.7\%&
98.0\%&
98.1\%&
98.2\%\\

\end{tabular}
\label{Lcheckerboard2}
\end {table}

\begin{table}[h!]
  \center
  \caption {Checkerboard Accuracies with 4 Hidden Layers. Compare to Table~\ref{checkerboard4}.}
  \footnotesize
\begin{tabular}{r||c|c|c|c|c}
  \toprule
  & \multicolumn{5}{c}{Hidden Units Per Layer}\\
  Activation & 5 & 10 & 50 &100 & 200\\
  \midrule
  \midrule  
 
tanh
&
95.0\%&
97.7\%&
98.4\%&
98.5\%&
98.1\%\\
relu
&
84.8\%&
96.7\%&
97.9\%&
97.8\%&
97.9\%\\
\midrule
r-sudo-2
&
52.1\%&
52.6\%&
52.0\%&
52.6\%&
55.0\%\\
r-sudo-4
&
53.1\%&
57.2\%&
96.6\%&
97.7\%&
98.4\%\\
r-sudo-8
&
61.4\%&
93.1\%&
97.7\%&
96.7\%&
97.7\%\\
r-sudo-16
&
72.2\%&
93.0\%&
98.3\%&
98.1\%&
98.1\%\\
r-sudo-32
&
80.2\%&
93.8\%&
98.2\%&
97.9\%&
97.7\%\\
r-sudo-64
&
80.2\%&
95.4\%&
98.1\%&
98.0\%&
97.8\%\\
r-sudo-128
&
84.7\%&
95.2\%&
98.2\%&
98.1\%&
98.0\%\\
r-sudo-256
&
85.1\%&
95.8\%&
98.1\%&
98.1\%&
98.0\%\\

\end{tabular}
\label{Lcheckerboard4}
\end {table}

\clearpage
\subsection {Simple Regression with R-SUDO units}

\begin{table}[h]
  \center
  \caption {Error in approximating a real-value function.  Networks have 2  hidden layers.  Compare to Table~\ref{approximate2}.}
    \footnotesize

\begin{tabular}{r||c|c|c}
  \toprule
  & \multicolumn{3}{c}{Hidden Units Per Layer}\\
  Activation & 10 & 20 & 50 \\
  \midrule
  \midrule

tanh
&
0.55&
0.24&
0.11\\
relu
&
3.13&
0.36&
0.13\\
\midrule
r-sudo-2
&
21.45&
21.47&
15.53\\
r-sudo-4
&
14.20&
6.85&
2.08\\
r-sudo-8
&
7.64&
1.89&
0.70\\
r-sudo-16
&
3.94&
0.90&
0.29\\
r-sudo-32
&
2.60&
0.53&
0.14\\
r-sudo-64
&
1.91&
0.39&
0.09\\
r-sudo-128
&
1.78&
0.31&
0.07\\
r-sudo-256
&
1.31&
0.26&
0.06\\

\end{tabular}
\label{Lapproximate2}
\end {table}

\begin{table}[h]
  \center
  \caption {Error in approximating a real-value function.  Networks have 4   hidden layers. Compare to Table~\ref{approximate4}.}
    \footnotesize

\begin{tabular}{r||c|c|c}
  \toprule
  & \multicolumn{3}{c}{Hidden Units Per Layer}\\
  Activation & 10 & 20 & 50 \\
  \midrule
  \midrule

tanh
&
0.13&
0.02&
0.01\\
relu
&
0.47&
0.11&
0.02\\
\midrule
r-sudo-2
&
21.40&
21.74&
19.97\\
r-sudo-4
&
16.26&
10.75&
3.22\\
r-sudo-8
&
9.16&
2.19&
0.55\\
r-sudo-16
&
3.26&
0.78&
0.26\\
r-sudo-32
&
1.90&
0.36&
0.10\\
r-sudo-64
&
0.81&
0.20&
0.05\\
r-sudo-128
&
0.44&
0.10&
0.03\\
r-sudo-256
&
0.41&
0.07&
0.02\\

\end{tabular}
\label{Lapproximate4}
\end {table}

\clearpage
\subsection {Memorization with R-SUDO units}
\label{rmemorization}
\begin{table}[!h]
  \center
  \caption {SSE memorizing image.  Networks have 2   hidden layer. Last Activation is tanh. Compare to Table~\ref{memorize2}.}
    \footnotesize

\begin{tabular}{r||c|c|c}
  \toprule
  & \multicolumn{3}{c}{Hidden Units Per Layer}\\
  Activation & 50 & 100 & 200 \\
  \midrule
  \midrule

tanh
&
1276.81&
1318.71&
1350.49\\
relu
&
1187.68&
1168.15&
1129.43\\
\midrule
r-sudo-2
&
1603.53&
1521.72&
1443.99\\
r-sudo-4
&
1372.24&
1314.86&
1299.39\\
r-sudo-8
&
1231.86&
1184.61&
1175.17\\
r-sudo-16
&
1171.52&
1129.30&
1105.97\\
r-sudo-32
&
1144.96&
1117.43&
1095.69\\
r-sudo-64
&
1133.57&
1100.98&
1094.33\\
r-sudo-128
&
1113.24&
1101.18&
1089.25\\
r-sudo-256
&
1118.84&
1115.16&
1079.66\\

\end{tabular}
\label{Lmemorize2}
\end {table}

\begin{table}[!h]
  \center
  \caption {SSE memorizing image.  Networks have 4 hidden layers. Last Activation is tanh. Compare to Table~\ref{memorize4}.}
    \footnotesize

\begin{tabular}{r||c|c|c}
  \toprule
  & \multicolumn{3}{c}{Hidden Units Per Layer}\\
  Activation & 50 & 100 & 200 \\
  \midrule
  \midrule

tanh
&
992.67&
978.79&
995.11\\
relu
&
915.97&
879.11&
846.82\\
\midrule
r-sudo-2
&
1661.37&
1633.06&
2442.41\\
r-sudo-4
&
1364.00&
1282.73&
1215.14\\
r-sudo-8
&
1185.38&
1109.68&
1070.57\\
r-sudo-16
&
1054.54&
995.39&
930.72\\
r-sudo-32
&
959.88&
903.00&
854.91\\
r-sudo-64
&
924.95&
871.47&
812.99\\
r-sudo-128
&
891.32&
848.97&
801.34\\
r-sudo-256
&
897.88&
815.65&
802.37\\

\end{tabular}
\label{Lmemorize4}
\end {table}

\begin{table}[!h]
  \center
  \caption {SSE memorizing image.  Networks have 10   hidden layers. Last Activation is tanh.Compare to Table~\ref{memorize10}.}
    \footnotesize

\begin{tabular}{r||c|c|c}
  \toprule
  & \multicolumn{3}{c}{Hidden Units Per Layer}\\
  Activation & 50 & 100 & 200 \\
  \midrule
  \midrule

tanh
&
898.96&
909.04&
875.31\\
relu
&
751.49&
701.92&
681.25\\
\midrule
r-sudo-2
&
1771.92&
1775.78&
1758.56\\
r-sudo-4
&
1552.20&
1505.91&
1432.91\\
r-sudo-8
&
1287.86&
1199.02&
1145.87\\
r-sudo-16
&
1097.75&
1001.96&
950.51\\
r-sudo-32
&
920.75&
868.79&
800.57\\
r-sudo-64
&
838.35&
779.91&
729.90\\
r-sudo-128
&
795.00&
736.38&
740.50\\
r-sudo-256
&
811.02&
674.58&
680.06\\

\end{tabular}
\label{Lemorize10}
\end {table}

\clearpage
\subsection {Autoencoding with R-SUDO units}

In Section~\ref{autoencoding}, these errors were presented relative to
the average performance of the 5 trials with the tanh units in a
convolution network.  That same baseline is used here to make the
comparison straight-forward.

\begin{table}[!h]
  ~
  \center
  \caption {Autoencoding Networks Convolution-Based - Relative SSE Errors. Compare to Table~\ref{autoConv}.}
  \footnotesize
\begin{tabular}{r||c|c|c|c}
  \toprule
  & \multicolumn{4}{c}{Network Scale}\\
  Activation & 1 & $\times2$ & $\times4$ & $\times8$\\  
  \midrule
  \midrule  
 
tanh
&
1.00&
0.68&
0.36&
0.06\\
relu
&
1.02&
0.68&
0.39&
0.08\\
\midrule
r-sudo-2
&
5.08&
5.11&
4.79&
2.53\\
r-sudo-4
&
1.96&
1.46&
1.05&
0.71\\
r-sudo-8
&
1.30&
0.97&
0.66&
0.35\\
r-sudo-16
&
1.14&
0.81&
0.48&
0.21\\
r-sudo-32
&
1.05&
0.73&
0.41&
0.14\\
r-sudo-64
&
1.01&
0.69&
0.38&
0.11\\
r-sudo-128
&
1.01&
0.67&
0.36&
0.10\\
r-sudo-256
&
1.05&
0.68&
0.35&
0.10\\

\end{tabular}
\label{LautoConv}
\end {table}

\begin{table}[h]
  ~  
  \center  
  \caption {Autoencoding Networks, Fully Connected - Relative SSE Errors. Compare to Table~\ref{autoFull}.}
  \footnotesize
\begin{tabular}{r||c|c|c|c}
  \toprule
  & \multicolumn{4}{c}{Network Scale}\\
  Activation & 1 & $\times2$ & $\times4$ & $\times8$\\    
  \midrule
  \midrule  
 
tanh
&
2.30&
2.00&
1.64&
1.30\\
relu
&
2.39&
1.99&
1.85&
1.47\\
\midrule
r-sudo-2
&
4.13&
3.82&
3.64&
3.53\\
r-sudo-4
&
3.54&
3.09&
2.77&
2.61\\
r-sudo-8
&
3.16&
2.74&
2.55&
2.25\\
r-sudo-16
&
2.97&
2.62&
2.38&
2.05\\
r-sudo-32
&
2.91&
2.58&
2.27&
1.96\\
r-sudo-64
&
3.05&
2.51&
2.12&
1.90\\
r-sudo-128
&
2.70&
2.40&
2.16&
1.83\\
r-sudo-256
&
2.73&
2.44&
2.14&
1.82\\

\end{tabular}
\label{LautoFull}
\end {table}

\clearpage
\subsection {MNIST with R-SUDO units}

\begin{table}[!h]
  \center
  \caption {MNIST Accuracy Results.  Networks have 1 hidden layer. Compare to Table~\ref{mnist1}.}
  \footnotesize
\begin{tabular}{r||c|c|c|c|c|c}
  \toprule
  & \multicolumn{6}{c}{Hidden Units Per Layer}\\
  Activation & 2 & 3 &4 & 10 & 50 & 100\\
  \midrule
  \midrule
 
tanh
&
63.9\%&
78.8\%&
85.9\%&
93.6\%&
97.1\%&
97.7\%\\
relu
&
66.1\%&
80.9\%&
86.5\%&
93.8\%&
97.2\%&
97.9\%\\
\midrule
r-sudo-2
&
27.6\%&
36.1\%&
42.3\%&
76.7\%&
96.3\%&
97.2\%\\
r-sudo-4
&
43.5\%&
64.2\%&
80.8\%&
91.8\%&
96.4\%&
97.3\%\\
r-sudo-8
&
48.2\%&
72.8\%&
82.1\%&
92.3\%&
96.6\%&
97.5\%\\
r-sudo-16
&
52.1\%&
73.8\%&
84.0\%&
92.4\%&
96.6\%&
97.3\%\\
r-sudo-32
&
55.1\%&
72.4\%&
83.0\%&
92.7\%&
96.6\%&
97.4\%\\
r-sudo-64
&
53.6\%&
74.4\%&
84.9\%&
92.7\%&
96.6\%&
97.4\%\\
r-sudo-128
&
51.3\%&
74.6\%&
84.0\%&
92.6\%&
96.7\%&
97.5\%\\
r-sudo-256
&
52.1\%&
72.3\%&
83.3\%&
92.7\%&
96.6\%&
97.4\%\\

\end{tabular}
\label{Lmnist1}
\end {table}

\begin{table}[!h]
  \center  
  \caption {MNIST Accuracy Results.  Networks have 4 hidden layers. Compare to Table~\ref{mnist4}.}
  \footnotesize

\begin{tabular}{r||c|c|c|c|c|c}  
  \toprule
  & \multicolumn{6}{c}{Hidden Units Per Layer}\\
  Activation & 2 &3 &4 & 10& 50& 100\\  
  \midrule
  \midrule
 
tanh
&
63.8\%&
73.5\%&
83.8\%&
93.9\%&
97.3\%&
97.9\%\\
relu
&
30.3\%&
54.1\%&
81.2\%&
93.7\%&
97.2\%&
97.8\%\\
\midrule
r-sudo-2
&
15.0\%&
17.8\%&
25.0\%&
47.2\%&
91.6\%&
96.4\%\\
r-sudo-4
&
18.7\%&
17.3\%&
33.0\%&
90.4\%&
96.6\%&
97.6\%\\
r-sudo-8
&
14.1\%&
28.6\%&
58.2\%&
92.1\%&
96.8\%&
97.7\%\\
r-sudo-16
&
25.2\%&
46.8\%&
56.4\%&
92.2\%&
96.9\%&
97.5\%\\
r-sudo-32
&
25.4\%&
58.4\%&
54.6\%&
92.7\%&
97.0\%&
97.7\%\\
r-sudo-64
&
29.9\%&
43.3\%&
71.5\%&
92.5\%&
96.9\%&
97.5\%\\
r-sudo-128
&
33.2\%&
56.9\%&
63.0\%&
92.7\%&
96.9\%&
97.7\%\\
r-sudo-256
&
31.1\%&
41.0\%&
63.9\%&
92.5\%&
96.8\%&
97.8\%\\

\end{tabular}
\label{Lmnist4}
\end {table}
\end{appendices}

\end{document}